# Estimating Link Flows in Road Networks with Synthetic Trajectory Data Generation: Reinforcement Learning-based Approaches

Miner Zhong
The University of Queensland
m.zhong@uq.edu.au

Jiwon Kim
The University of Queensland
jiwon.kim@uq.edu.au

Zuduo Zheng
The University of Queensland
zuduo.zheng@uq.edu.au

**Abstract:**

This paper addresses the problem of estimating link flows in a road network by combining limited traffic volume and vehicle trajectory data. While traffic volume data from loop detectors have been the common data source for link flow estimation, the detectors only cover a subset of links. Vehicle trajectory data collected from vehicle tracking sensors are also incorporated these days. However, trajectory data are often sparse in that the observed trajectories only represent a small subset of the whole population, where the exact sampling rate is unknown and may vary over space and time. This study proposes a novel generative modelling framework, where we formulate the link-to-link movements of a vehicle as a sequential decision-making problem using the Markov Decision Process framework and train an agent to make sequential decisions to generate realistic synthetic vehicle trajectories. We use Reinforcement Learning (RL)-based methods to find the best behaviour of the agent, based on which synthetic population vehicle trajectories can be generated to estimate link flows across the whole network. To ensure the generated population vehicle trajectories are consistent with the observed traffic volume and trajectory data, two methods based on Inverse Reinforcement Learning and Constrained Reinforcement Learning are proposed. The proposed generative modelling framework solved by either of these RL-based methods is validated by solving the link flow estimation problem in a real road network. Additionally, we perform comprehensive experiments to compare the performance with two existing methods. The results show that the proposed framework has higher estimation accuracy and robustness under realistic scenarios where certain behavioural assumptions about drivers are not met or the network coverage and penetration rate of trajectory data are low.

## 1. Introduction

The accurate estimation of traffic flows (or volumes) on road links is critical in managing a road network and evaluating its performance. While loop detectors are installed to collect link flow data, the observation points are often limited to a subset of links and there are still a large proportion of links that do not have direct observations. Thus, unobserved link flows need to be estimated based on available data and this is referred to as the link flow estimation problem in the transportation literature (Abadi et al. 2015; Brunauer et al., 2017; Lederman and Wynter, 2011; Owais et al. 2020; Van Oijen et al. 2020).

Recent years have witnessed an increasing application of vehicle detection technologies on the road network (e.g., GPS, WiFi, Bluetooth, and RFID devices). A vehicle trajectory is a time-ordered sequence of locations visited by the vehicle (in latitudes and longitudes). A collection of vehicle trajectories usually offers deep insights into vehicle propagation information. Since the entire travel paths of vehicles are captured, trajectory data have better spatial coverage than traffic volume data from loop detectors. It is thus desirable to combine these two data sources to improve link flow estimation. However, considering the limited market penetration rate of vehicle detection technologies and the data collection errors, the observed vehicle trajectory data may not be representative of the true population trajectory distribution. Link flows estimated from these trajectory data may not be concordant with those estimated from the traffic volume data. Thus, the key question becomes how to infer the traffic flow scenario that describes the true population vehicle movements from these two limited data sources.

Most previous studies that attempted to incorporate trajectory data in link flow estimation make strong assumptions about the observed trajectory data. For example, Brunauer et al. (2017) proposed a method to propagate the observed link flows to unobserved links using propagation rules defined based on the vehicle trajectory data. They assume that such trajectories cover most of the link-to-link connections in the road network. In the transportation literature, the link flow estimation problem is often solved within an origin-destination (OD) estimation problem, where trajectory data are used to estimate either link flow distribution (Guo et al., 2019; Ma and Qian, 2018a; Yang et al., 2017) or path flow distribution (Carrese et al., 2017; Nigro et al., 2018; Parry and Hazelton, 2012). These studies usually assume that the market penetration rate of vehicle tracking devices is fixed and uniform across the network. Such assumptions made in the existing studies, however, are often violated in real-world trajectory datasets.

In this paper, we propose methods to estimate link flows for a given time interval from limited traffic volume data and sparse trajectory data with minimal assumptions and requirements on the available data. Specifically, we aim to address the following challenges:

- Traffic volume data capture the traffic flow of the whole population at observation points. However, the spatial coverage of the available volume data is limited because only a small subset of links has detectors installed.
- Trajectory data capture the vehicles' route preferences and movement patterns across the network. In this paper, only the spatial aspect of vehicle trajectories will be considered. That is, each observed trajectory is translated into a time-ordered sequence of links. The available trajectory data are sparse in that they only represent a sample of the whole population. The percentage of vehicle trajectories being captured in the observed trajectory set (hereafter referred to as the sampling rate) is unknown. The distribution of such sampling rates across the network is not uniform, meaning that some parts of the network may have a higher percentage of vehicles being detected than other parts of the network do. Also, some paths and links on the network may not be covered by the observed trajectories.

Figure 1 shows an illustrative example of the proposed research problem. A road network is represented as a directed graph, where edges are road links and nodes are intersections. The top two figures show the true path flows and true link flows that are unknown to a modeller, respectively. The numbers on the figures represent the path and link flows in this road network. Four different paths are used by vehicles, represented as red, green, blue, and black paths on the top-left figure, respectively. Consider the scenario where the modeller can observe partial path flows from trajectory data and a subset of link flows from traffic volume data, as shown in the first two figures at the bottom. More specifically, the available vehicle trajectory data cover three different paths (in red, green, and blue, respectively), while no trajectory is observed on the black path (bottom-left figure). The trajectory sampling rates are different across the paths. For instance, 20% of vehicles along the red path are captured by the trajectory data, while 50% of vehicles are captured along the green path. These sampling rates are not known to the modeller. The traffic volume data are available only on five links (bottom-middle figure). With these observed data, the modeller' goal is to estimate link flows across the whole network, as shown in the bottom-right figure.

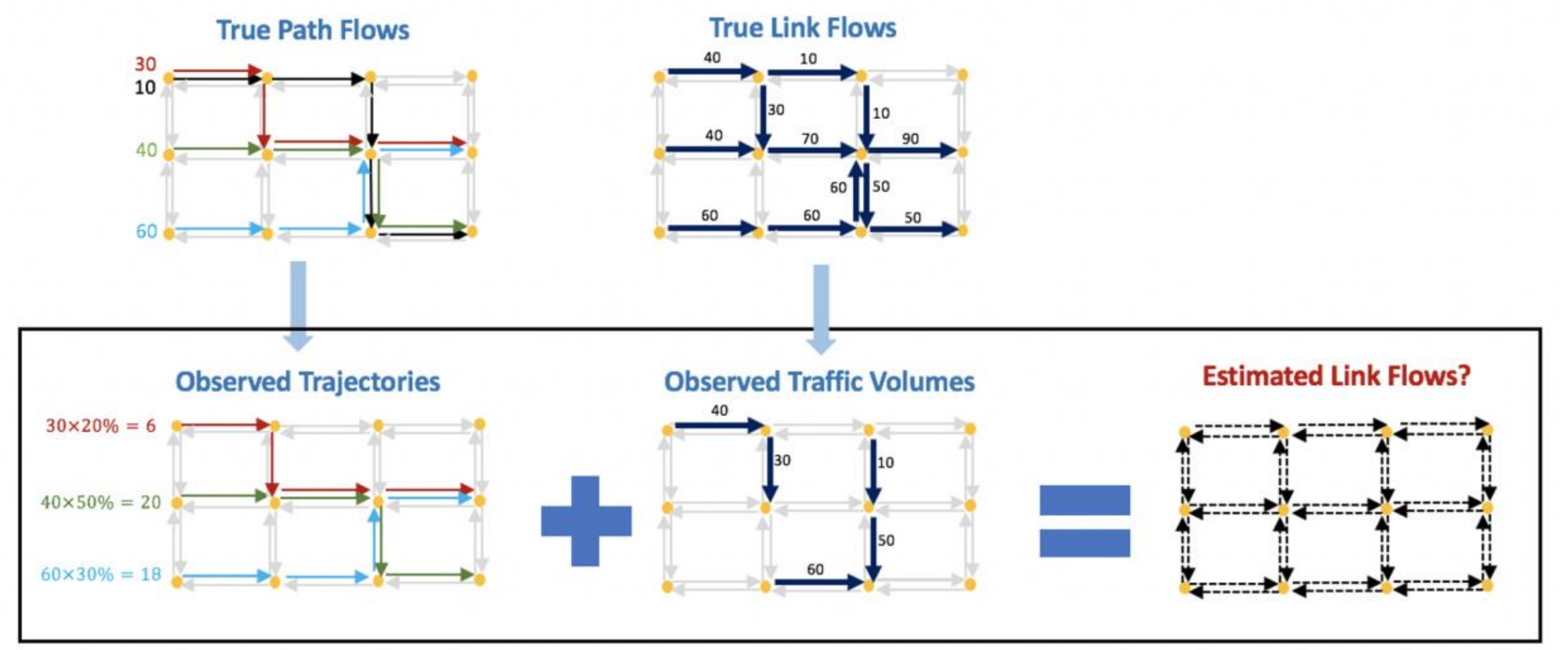

Figure 1. An illustrative example of the link flow estimation problem considered in this study

Note that the targeted problem is underdetermined. There may exist multiple vehicle movement scenarios that can produce the same data observations. Instead of finding the true vehicle movement scenario, which requires additional data and/or assumptions, the focus of this paper is to find a plausible scenario that provides reasonable estimates of the unobserved link flows based only on the limited observed data. To identify a plausible scenario in a principled way, we develop a *generative modelling* approach based on the Markov Decision Process (MDP) framework. Generative models aim to learn the underlying distribution of the observed data samples and generate new data points with some variations, namely, to generate *synthetic data*. Recently, synthetic data generation has been widely used as a way to increase the data sample size when data collection is expensive or to provide alternative representations when the original data cannot be directly used for privacy protection reasons (Bindschaedler and Shokri, 2016; Martinsson et al., 2018; Yin et al., 2018). In our problem, the vehicle trajectory data can be viewed as data samples from the true population trajectory distribution. We aim to design a generative modelling framework to learn a population trajectory distribution from the observed data. Once a population trajectory distribution is obtained, it is possible to generate synthetic trajectories to represent the whole vehicle population such that traffic volumes on any unobserved links can be estimated.

By representing a vehicle trajectory in terms of a sequence of links visited by a vehicle, each trajectory can be viewed as a result of a *sequential decision-making process* from the origin to the destination, where a decision is made at each link to determine the next link among possible downstream links. If we can model this underlying sequential decision-making process, a generative model can be constructed to generate synthetic population trajectory data that solve our research problem discussed above. Motivated by this view, we formulate the trajectory data generation procedure as an MDP, which is a classical formalisation of sequential decision making (Bachman and Precup, 2015; Sutton and Barto, 2018). In an MDP, there is a decision-making *agent*, which interacts with its *environment* over a sequence of discrete time-steps. At each step, the agent encounters a *state* and chooses an *action*. The environment then responds by moving to a new state and giving the agent a *reward*. The agent's objective is to maximise the amount of reward it receives over time. A *policy* is a function specifying which action the agent will choose at each state. In our problem, which we refer to as *road-network MDP*, each road link is considered a state and the transitions from one link to the next link are considered actions. The policy is then a function describing a probability distribution over link transitions, which is essentially a probability distribution of vehicle trajectories. The goal of our generative modelling approach is thus to first find a policy in the road-network MDP and then use this policy to generate synthetic vehicle trajectories for link flow estimation. In this approach, the main challenge is how to find a policy that can capture the patterns in the given trajectory data samples while ensuring that the generated population trajectories satisfy the constraints imposed by the given traffic volume data.

Reinforcement Learning (RL) is a powerful way to learn policies for sequential decision-making tasks in MDPs (Sutton and Barto, 2018). Standard RL models aim to find a policy that maximizes the cumulative rewards of an agent's decisions, by allowing the agent to explore its environment and giving the agent feedback about its actions based on a reward function. However, in the road-network MDP, the reward function is not known. More importantly, the constraints that we want to impose (i.e., the generated trajectories should be consistent with the link traffic volume data) cannot be expressed as standard reward functions. Some variations of RL models can deal with such cases where defining a reward function to express the desired behaviour is challenging. One approach is *Inverse Reinforcement Learning* (IRL), which learns a reward function from data, instead of specifying it manually (Ziebart et al., 2008a). Another approach is reinforcement learning under constraints, which we refer to as *Constrained Reinforcement Learning* (CRL). The CRL defines the learning goal in terms of vector-valued constraints, instead of a scalar reward (Miryoosefi et al., 2019). Motivated by these previous approaches, this paper proposes two RL-based methods that can be implemented within the proposed generative modelling framework, namely: (1) Inverse Reinforcement Learning for link Flow estimation (IRL-F) and (2) Constrained Reinforcement Learning for link Flow estimation (CRL-F). By using either IRL-F or CRL-F, we can find a policy in the road-network MDP that allows the generated vehicle trajectories to be consistent with the observed data. Once this policy is found, the next question is to determine the number of synthetic vehicle trajectories such that link flow estimates on unobserved links can be obtained in terms of the number of times a given link is visited by the synthetic trajectories. This paper develops a simple technique to determine the optimal size of the synthetic trajectory set for solving the link flow estimation problem. The contributions of this paper are summarised as follows:

- The proposed generative modelling framework makes it possible to estimate link flows across the network without relying on expensive infrastructures such as loop detectors covering every link in the road network.

- The proposed framework allows the incorporation of trajectory data in the link flow estimation problem, where the assumptions that the observed trajectories have known and uniform sampling rates and/or cover most link-to-link transitions in the road network are all relaxed.
- The proposed framework provides a data-driven solution that avoids assumptions about travellers' route choice behaviour in the road network. Instead, the observed traffic data are used to recreate the underlying vehicle movement scenarios and generate synthetic trajectories. To the best of our knowledge, this is the first work to solve the link flow estimation problem using the concept of synthetic trajectory data generation.
- The proposed IRL-F and CRL-F methods are novel extensions to the existing RL methods in the literature, both of which can be used to solve the road-network MDP without requiring knowledge about reward functions.

The paper is structured as follows: Section 2 provides a literature review on the relevant research topics. Section 3 describes the proposed generative modelling framework for link flow estimation. Section 4 provides experiment results to validate the proposed framework and Section 5 presents conclusions.

## 2. Literature review

This section first provides a review of previous link flow estimation methods, followed by a review of the applications of generative models and RL methods in the transportation literature that are relevant to the research topic addressed in this paper.

### 2.1 Link flow estimation

Most link flow estimation methods primarily consider the traffic volume data collected by loop detectors. Some previous studies aim to find the smallest subset of detectors that enables accurate link flow estimation across the network. Hu et al. (2009) formalized this problem from the budgetary planning perspective. The results are the optimal locations for passive link flow counting sensors. Castillo et al. (2011) proposed another method considering both the passive counting sensors and plate scanners. However, in most situations, the link flow estimation problem must be solved with traffic volume data collected from sensors that have already been installed in non-optimal locations in the road network. Lederman and Wynter (2011) proposed a two-phase solution framework. In the offline phase, the link-to-link splitting probabilities are determined according to traffic equilibrium principles. These probabilities are used in the online phase to propagate the observed link flows to unobserved links. Abadi et al. (2015) used the traffic volume data as inputs to a traffic simulator together with other information about the traffic status, the estimation results are obtained from simulations. Morimura et al. (2013) formulated the link flow estimation problem as an optimization problem based on a partially observed Markov Chain. This formulation requires assumptions about the travellers' route choice behaviours. Ide et al. (2017) applied the same formulation in a real-life case where the traffic volume data are collected from traffic cameras. Only a few previous studies considered data fusion when solving the link flow estimation problem. Brunauer et al. (2017) proposed to solve a local network propagation problem between observed links based on propagation rules indicated by the probe vehicle trajectories. Such trajectories are assumed to cover most of the link-to-link transitions in the road network. Michau et al. (2017) proposed an estimation method for the link-based OD matrix, which is a disaggregated form of OD matrix in the form of link flows. Vehicle trajectory data are used in this method, with sampling rates assumed to be a single numerical value for each OD pair in the road network.

Many previous link flow estimation studies are coupled with studies on the OD estimation problem. Yang et al. (1992) formulated the OD estimation problem using a bi-level optimization model, where the upper-level model is a generalized least squares model aiming to minimize the deviation between the observed and the estimated traffic data while the lower-level model is a traffic assignment problem that establishes connections between OD flows and link flows. Nie and Zhang (2008) formulated the OD estimation problem as a variational inequality, where the assignment matrix is endogenized. Shen and Wynter (2012) proposed a one-level convex optimization model, which can be viewed as a special case of the traffic assignment problem with elastic demand. Most OD estimation methods require route choice assumptions to solve the traffic assignment problem. Static user equilibrium has been commonly assumed (Lundgren and Peterson, 2008; Yang et al., 1992). Some recent studies have made more complicated equilibrium assumptions. For example, Ma and Qian (2018b) utilized the stochastic user equilibrium assumptions, which do not require travellers to have complete knowledge of the network conditions. Notably, some recent studies proposed to use the observed traffic data to replace the traffic assignment procedure. Kim and Jayakrishnan (2010) proposed to compute the traffic assignment ratios from the probe vehicle data, which are assumed to be able to reveal the true traffic flow pattern. Similar assumptions were made in Vogt et al. (2019), where the observed trajectory data are used to calculate the number of turns at each intersection in the road network. The OD estimation problem has also been cast in statistical models. Parry and Hazelton (2012) proposed a likelihood-based inference model based on traffic volume data and sporadic vehicle routing data, assuming the vehicle tracking probability is a fixed number across the network. Castillo et al. (2013) proposed a conjugate Bayesian method to estimate OD demand, path flows and link flows at the same time. This method requires sufficient plate scanners to be installed in the road network to provide plate scanning data for the traffic flow estimation.

Another relevant study to the proposed link flow estimation problem is the link utilization method discussed in Van Oijen et al. (2020). The authors proposed to first estimate the parameters for the recursive logit model proposed by Fosgerau et al. (2013) using routing data collected from a set of fixed proximity sensors. The estimated recursive logit model is then used to provide link utilization information on networks. The recursive logit model shares some similarities with the generative model discussed in this paper, as both models aim to learn the sequential decision rules underlying the observed data. However, there are clear differences as follows: (1) The recursive logit model aims to assign path probabilities in a way that is consistent with rational route choice behaviour, which requires a utility function described using road characteristics chosen by domain experts. In contrast, the path probabilities in the proposed generative model are determined to satisfy the constraints imposed by the observed data. There is no assumption on route choice behaviours. (2) Only one type of observed data is discussed in Van Oijen et al. (2020) under certain requirements on network configurations. In contrast, the proposed generative method considers two types of observed data with minimal assumption on data availability. (3) The link utilization results obtained by Van Oijen et al. (2020) can only be interpreted in a relative way. In contrast, the link flow estimation results provided by the proposed generative method are absolute values, which are of greater importance from the perspective of network management.

In sum, the previous link flow estimation studies have identified the potential of using vehicle trajectory data, but often made strong assumptions about the data coverage. Also, many studies require assumptions about route choice behaviours and/or equilibrium conditions on the road network. However, there may not exist sufficient evidence to support such route choice behaviour assumptions or determine which equilibrium can best describe real-life situations.

### 2.2 Generative models

Generative models focus on finding the underlying distribution given some observed data samples and utilize this distribution to generate new data points. Recent years have seen many applications of such models in the transportation field. One prominent application is in synthetizing mobility sequences.

Bindschaedler and Shokri (2016) proposed a generative model to synthesize individual geolocation traces based on location visitation sequences from a real mobility dataset. Synthetic data can protect the privacy of real mobility data. Yin et al. (2018) proposed a model for generating human travel itineraries. The authors annotated the mobile phone call detail records using activities and create an input-output hidden Markov model. The synthetic itineraries can replace itinerary data collected in surveys. Chen et al. (2019) proposed to learn the latent factors behind the observed human travel trajectories. The learned latent factors are used to build a generative model for trajectory synthesis. Generative models have also been discussed in the field of vehicle manoeuvre planning. For example, Wheeler et al. (2015) proposed a generative model to learn the evolution pattern of traffic scenes on highways from real testing data. The proposed model can generate synthetic traffic scene propagations, which are used to train autonomous vehicles. Krajewski et al. (2019) proposed to use a Variational Autoencoder to learn the pattern of the vehicle manoeuvre trajectories from real data and generate synthetic trajectories for training purposes. Overall, various generative models have been proposed to solve transportation problems. However, there is a lack of research in generating synthetic data by combining multiple data sources and the application of this generative modelling or synthetic data generation approach to traffic flow estimation problems has not been studied.

### 2.3 Reinforcement Learning methods

The applications of different RL methods in the transportation literature depends on the availability of reward functions. Standard RL methods require reward functions to be known. A typical application of such methods is to solve traffic control problems. For example, Zhu and Ukkusuri (2014) proposed to determine speed limits by applying a standard RL method. The speed limit controller is designed as an agent. The rewards are designed to achieve certain traffic planning objectives. Hoel et al. (2018) proposed an RL method for vehicle manoeuvre planning, where the agent represents a vehicle. The agent is trained with manually designed rewards. For situations where it is challenging to manually determine reward functions, inverse reinforcement learning (IRL) has been applied. For example, Ziebart et al. (2008a) proposed a method called Maximum Entropy IRL (MaxEnt IRL), where the route choice decision of drivers are modelled in an MDP. The goal is to recover a reward function by viewing the available vehicle trajectory data as expert's demonstrations. The same authors extend the MaxEnt IRL to learn the context-based routing preferences in the following paper (Ziebart et al., 2008b). Choi and Kim (2015) proposed a hierarchical Bayesian IRL framework to address the suboptimality in the expert's demonstrations. This framework was applied to infer drivers' routing preferences given GPS trajectories. Pan et al. (2019) proposed to learn the taxi drivers' preferences towards different regions using the Relative Entropy IRL method. The taxi drivers' utility functions are recovered based on the observed trajectory data. In summary, RL methods have been used to solve a wide range of transportation problems. In particular, many studies leveraged IRL methods to learn driver's routing preferences from observed trajectory data, which are especially relevant to our study. There are, however, research gaps that the previous studies have not attempted to learn driving patterns from multiple data sources. For instance, learning vehicle trajectories by considering both sample trajectory and traffic volume data has not been studied. In this paper, we address these limitations identified in the existing literature by developing data-driven methods to leverage the available traffic volume and trajectory data to generate synthetic trajectories and solve link flow estimation problems.

## 3. Methodology

This section first gives an overview of the proposed generative modelling framework and then describes how RL-based methods (i.e., IRL-F and CRL-F) can be applied in this framework to solve the link flow estimation problem.

### 3.1 The modelling framework based on MDP

A generative modelling framework is proposed to generate synthetic population trajectories, which are then used to estimate unobserved link flows in a road network. Figure 2 shows the overall concept of the proposed framework. Given link flow data from a subset of links and vehicle trajectory data from a subset of vehicles, the proposed IRL-F and CRL-F models learn a policy of a road-network MDP that mimics the true population trajectory distribution underlying the observed traffic data (process 1 in Figure 2). The most straightforward approach to estimating link flows using the learned policy is to generate synthetic vehicle trajectories for the whole population by sampling from the policy and counting the number of trajectories passing each link (process $2'$). This approach is, however, computationally expensive due to a large number of trajectories that need to be generated until a proper population size is found. Instead, a simple approach is proposed to using *state visitation frequencies* measuring how often each link is visited by trajectories, which can be calculated as a by-product of the training process of the IRL-F and CRL-F models (process 2). The link flows for all links can be estimated based on the state visitation frequencies from the road-network MDP and the link flows for the observed links can be validated against the actual volume data (process 3). The detailed methodologies for each process implemented in this generative modelling framework are discussed in the rest of this section.

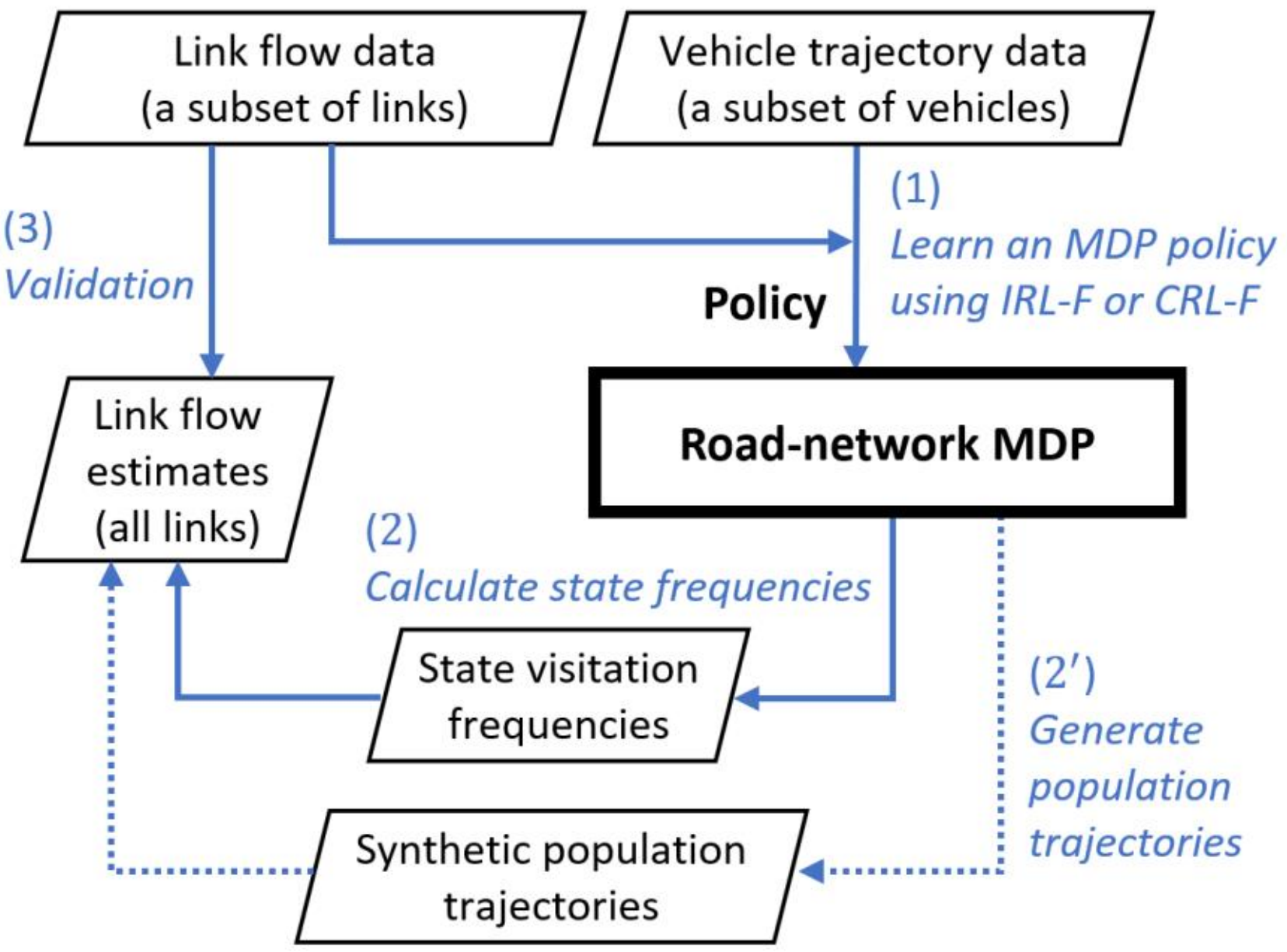

Figure 2. The proposed MDP-based generative modelling framework for link flow estimation

Vehicles' sequential decision-making to perform link-to-link transitions in a road network can be modelled using a finite-horizon episodic MDP with absorbing states. In an episodic MDP, the agent-environment interaction breaks down into a series of separate episodes (episodic tasks), each of which consists of a finite sequence of time steps, rather than one long sequence of time steps (continuing tasks). Vehicle trajectories are naturally expressed as episodic tasks, where each trajectory represents one episode in the MDP. This paper proposes to formulate the link flow estimation problem based on a road-network MDP, which can be described by a tuple $M = (S, A, \mu_0, P_T, r, \gamma, H)$:

- $S$ is the set of states, which includes all links in the road network as well as an additional set of virtual links representing absorbing states corresponding to the end of an episode. First, possible destination locations in the network are identified by extracting a subset of links where the observed vehicle trajectories ended. Then, a virtual link is added to each of these identified links to allow an action to terminate trips on those possible destination locations. The subset $S_v$ is defined as the subset of states that correspond to the links that have loop detectors installed (i.e., links that have available traffic volume count data).

- $A$ is the set of actions, which are possible transitions from one link to the next link in the road network.
- $\mu_0$ is the initial state distribution, which is a probability distribution over the set of links to start with. The initial state distribution is assumed to be equal to the distribution of initial links visited by the observed vehicle trajectories.
- $P_T: S \times S \times A \rightarrow [0, 1]$ is the state transition probability. $P_T(s_{t+1}|s_t, a_t)$ represents a probability of visiting $s_{t+1}$ in the next time-step by choosing action $a_t$ in state $s_t$ at the current time-step $t$. In the road-network MDP, the transition probability is known and deterministic in that choosing an action to move to a specific downstream link would indeed lead to that link with the probability of 1 and the probability of ending up in other links is zero.
- $r: S \times A \times S \rightarrow \mathbb{R}$ is the reward function, where $r_t = r(s_t, a_t, s_{t+1})$ is the reward associated with the transitioning to state $s_{t+1}$ in the next time step by choosing action $a_t$ in state $s_t$ at the current time step $t$. Such a reward function is not known.
- $\gamma \in [0, 1]$ is the discount factor, which shows how much future reward should be discounted when the agent is making decisions. It is assumed to be 0.99.
- $H$ is the horizon, which is the maximum number of steps in each episode. It is assumed to be equal to the maximum length of the observed vehicle trajectories.

A policy $\pi: S \rightarrow A$ is a function that maps a state to an action to take in that state, where $\pi(a|s)$ is the probability of selecting action $a$ in state $s$ under policy $\pi$. In the road-network MDP, this function represents the probability of a decision-making agent (vehicle) choosing the next link among a set of downstream links on the current link. For each episode, the agent starts from an initial state $s_0$. At each step $t = 0, 1, 2, \cdots, H$, the agent chooses an action $a_t$ given the current state $s_t$ based on the policy $\pi$, which results in transitioning to the next state $s_{t+1}$ and receiving a reward $r_t$. The sequence of states and actions visited by the RL agent during an episode is normally called a *trajectory* in the RL literature, but we will refer to it as a *state-action path* $\tau = (s_0, a_0, s_1, a_1, \dots, a_{H-1}, s_H)$ to distinguish it from a vehicle trajectory in a road network. This state-action path can be translated into a spatial vehicle trajectory in the road network as the sequence of states visited by the agent represents the time-ordered sequence of road links travelled by a vehicle. In this paper, only the spatial aspect of vehicle trajectories (location sequences) will be considered, without the temporal aspect (travel time between locations).

Each episode is associated with a *return*, which is defined as the sum of the discounted rewards the agent received over the episode's state-action path. Standard RL methods aim to determine the optimal policy by maximizing the expected return, which is the expectation of returns of all possible state-action paths under the policy. In this paper, to learn the underlying population trajectory distribution given the observed data samples (i.e., vehicle trajectory data and traffic volume data), we aim to find a policy in the road-network MDP that allows the agent to generate state-action paths that can be viewed as synthetic population vehicle trajectories which are consistent with the observed data samples. It is difficult to manually specify reward functions in the road-network MDP to achieve this goal. Instead, the observed traffic data offer information on the road-network MDP from a different aspect.

- The vehicle trajectory data consist of time-ordered sequences of links visited by the detected vehicles, which can be translated into sequences of states visited by the agent ordered by time-steps. By associating an action to each state, each vehicle trajectory can be translated into a state-action path in the road-network MDP. Let $T_{obs}$ denote the set of state-action paths translated from the observed trajectory data. By counting the number of times each state (link) is visited by these trajectories, we can obtain a set of *state visitation* frequencies over the state set $(S)$, $D_{obs} = \{D_s^{T_{obs}} | \forall s \in S\}$, where $D_s^{T_{obs}}$ is the visitation count on state $s$ based on trajectory set $T_{obs}$, i.e., $D_s^{T_{obs}} = \sum_{\tau \in T_{obs}} \sum_{s' \in \tau} \mathbf{1}_s(s')$ and $\mathbf{1}_s(s')$ is the indicator function that returns 1 if $s' = s$ and 0 if $s' \neq s$. Note that the observed trajectories only account for a subset of the population trajectories with non-uniform sampling rates (which are unknown). Thus, state visitation counts from the observed trajectories might have a different distribution from the state visitation counts calculated from the true population trajectories, and the relationship between these two distributions is unknown.

- The traffic volume data are available each link that has a loop detector installed, which can be translated into the number of times each state is visited by the agent. We can obtain a set of state visitation counts over the detector links $(S_v)$ from these volume data, $Q_{obs} = \{v_s | \forall s \in S_v\}$, where $v_s$ represents the traffic volume on state $s$. Note that visitation counts on these states are equal to the counts derived from the true population trajectories because loop detectors installed on these links are assumed to be able to capture all vehicles passing the links. However, since traffic volumes are only observed on a subset of links (states), visitation counts on states outside $S_v$ are not available.

Based on the above-mentioned conditions, the proposed research objectives can be expressed as finding a policy in the road-network MDP that generates state-action paths whose state visitation count distribution mimics the true state visitation count distribution implied by both types of observed data. These generated state-action paths represent synthetic population vehicle trajectories that can be used to estimate unobserved link flows by estimating the state visitation counts for the states that do not have detectors.

### 3.2 Link flow estimation with IRL-F

In the road-network MDP, the agent needs to be trained to generate state-action paths that are consistent with the state visitation count distributions implied by the observed traffic data. We adopt the Maximum Entropy Inverse Reinforcement Learning (MaxEnt IRL) method proposed by Ziebart et al. (2008a), which can train the agent to generate state-action paths that mimic a set of state-action paths demonstrated by an expert. However, the original MaxEnt IRL cannot be directly applied to solve the road-network MDP because it assumes that the expert's demonstrations are in the form of state-action paths, whereas in our case we need to represent the expert's demonstrations not only in terms of state-action paths (to account for trajectory data) but also in terms of state visit counts for a subset of states (to account for traffic volume data). As such, we propose a method called IRL-F that modifies MaxEnt IRL to solve the road-network MDP and further propose a simple link flow estimation method based on the policy found by IRL-F.

#### 3.2.1 Maximum Entropy Inverse Reinforcement Learning

In this sub-section, we briefly introduce the MaxEnt IRL algorithm by Ziebart et al. (2008a).

**Environment definition:** Given an MDP, MaxEnt IRL assumes that each state $s \in S$ is characterized by a feature vector $\mathbf{f}_s \in \mathbb{R}^k$, where $k$ is the feature dimension. A state-action path $\tau$ is a sequence of all states and actions encountered by this agent, which is characterized by a path feature vector $\mathbf{f}_\tau \in \mathbb{R}^k$ that is defined as the sum of all state feature vectors in this path.

$$\mathbf{f}_\tau = \sum_{s \in \tau} \mathbf{f}_s \tag{1}$$

The agent makes sequential decisions based on some unknown reward values. It is assumed that visiting any state $s \in S$ incurs a state reward value that is linear to the state feature vector, parametrized by unknown reward weights $\theta \in \mathbb{R}^k$. The reward value $R_\theta(\tau)$ for a given path $\tau$ can be represented as the sum of state rewards along that path.

$$R_\theta(\tau) = \sum\nolimits_{s \in \tau} \theta^T \mathbf{f}_s \tag{2}$$

MaxEnt IRL considers the distribution over the set of paths that this agent can take, aiming to find the path distribution that mimics the distribution indicated by the expert's demonstrations. For deterministic MDPs, the path distribution is parameterized by the reward weights $\theta$. Let $P(\tau|\theta)$ denote the probability of taking path $\tau$ given reward parameter $\theta$, which can be expressed as follows.

$$P(\tau|\theta) = \frac{e^{R_\theta(\tau)}}{\sum_{\tau'} e^{R_\theta(\tau')}} \tag{3}$$

**IRL objective:** The expert's behaviour is represented by a set of demonstrated paths ($T_e$). To train the agent to behave following such demonstrations, MaxEnt IRL aims to find the optimal reward weight ($\theta^*$) that maximizes the likelihood of the expert's demonstrated paths under the maximum entropy distribution, which is expressed as follows.

$$\theta^* = \underset{\theta}{\operatorname{argmax}}\, L = \underset{\theta}{\operatorname{argmax}} \sum_{\tau \in T_e} \log P(\tau|\theta) \tag{4}$$

The optimal reward weight can be obtained using a gradient descend method. The gradient can be calculated as follows, where $M$ represents the number of demonstrated paths and $D_s$ represents the expected state visitation frequency on state $s$.

$$\begin{aligned} \nabla_\theta L &= \frac{1}{M} \sum_{\tau \in T_e} \mathbf{f}_\tau - \sum_{s \in S} D_s \mathbf{f}_s \\ &= \mathbf{f}_{\text{expert}} - \mathbf{f}_{\text{policy}} \end{aligned} \tag{5}$$

The first part of this gradient can be viewed as the expectation of path feature vectors over the expert's demonstrated paths, denoted by the expert's feature expectation $\mathbf{f}_{\text{expert}} \in \mathbb{R}^k$. The second part of this gradient, denoted by the policy feature expectation $\mathbf{f}_{\text{policy}} \in \mathbb{R}^k$, can be viewed as the expectation of path feature vectors over a set of paths generated under the current policy, which is determined by the current reward weight $\theta$. In this way, the gradient represents the difference between $\mathbf{f}_{\text{expert}}$ and $\mathbf{f}_{\text{policy}}$. The objective of MaxEnt IRL is thus to find the policy that matches the policy feature expectation with the expert's feature expectation.

**Solution process:** MaxEnt IRL can be solved using a gradient-based method. Generally, the reward weight $\theta$ can be randomly initialized, and then updated iteratively based on the gradient calculated using Eq. (5). Expert's feature expectation and policy feature expectation are needed to calculate such a gradient. The solution process will stop when the optimal value is found. The output of MaxEnt IRL is the optimal reward weight, based on which the reward function in the MDP can be recovered. Following the policy implied by this reward function, the agent generates the state-action paths that closely mimic the expert's demonstrations.

### 3.2.2 IRL-F

To solve the proposed link flow estimation problem, we propose a new method, IRL-F, which is adapted from the MaxEnt IRL method. Specifically, instead of observing the ground truth population trajectory set, we are given two types of observed data (i.e., vehicle trajectory data and traffic volume data), each of which only reflects part of the true network flows. IRL-F is proposed to train the agent using these observed data so that the optimal policy generates a trajectory distribution that mimics the population trajectory distribution. To achieve this goal, we introduce adaptations to the MaxEnt IRL method in the environment definition (i.e., new state feature definition) as well as the solution process (i.e., new gradient calculation process).

**Environment definition:** In the road-network MDP, each state $s \in S$ is characterized by a feature vector $\mathbf{f}_s \in \mathbb{R}^{k_1+k_2}$, which is a concatenation of two feature vectors $\mathbf{f}_\mathrm{s}^{(1)}$ and $\mathbf{f}_\mathrm{s}^{(2)}$. The objective of IRL-F is to match the policy feature expectation to the expert's feature expectation. Both types of observed data can be viewed as expert demonstrations. Therefore, the first part of the state feature vector, $\mathbf{f}_s^{(1)}$, is designed to facilitate feature expectation matching regarding the observed trajectory data, while the second part of the state feature vector, $\mathbf{f}_s^{(2)}$, is designed to facilitate feature expectation matching regarding the observed traffic volume data.

$$\mathbf{f}_s = \left[\mathbf{f}_\mathrm{s}^{(1)^T}, \mathbf{f}_\mathrm{s}^{(2)^T}\right]^T, \qquad \mathbf{f}_s^{(1)} \in \mathbb{R}^{k_1}, \mathbf{f}_s^{(2)} \in \mathbb{R}^{k_2} \tag{6}$$

In the road-network MDP, each link is represented by a state. We proposed a new state feature definition that is different from the definition in MaxEnt IRL. In the original MaxEnt IRL paper (Ziebart et al., 2008a). the MaxEnt IRL method was applied in the task of learning drivers' route choice behaviours from GPS trajectory data, where road segments are modelled as states and the state feature vector is defined in terms of four different road characteristics that describe each road segment: namely, road type, speed, lanes, and transitions. While using such general road features is useful in learning and interpreting the agent' route choice behaviour as a function of road characteristics, since our primary goal is to solve the link flow estimation problem rather than to learn accurate driver behaviours, using a unique feature associated with each link can better suit our purposes. For instance, with a unique link identification (link ID) as a feature, the feature expectation matching between the agent' and expert's trajectories can directly lead to the matching of the state visitation frequency for an individual link, which helps the agent replicate the link visitation patterns (and thus the link flow patterns) in observed trajectory and traffic volume data.

As such, we propose the use of unique link IDs as feature vectors for $\mathbf{f}_s^{(1)}$ and $\mathbf{f}_s^{(2)}$ as follows.

- The state feature vector $\mathbf{f}_s^{(1)} \in \mathbb{R}^{k_1}$ is designed to convey information from the vehicle trajectory data (i.e., the distribution of state visitation counts over the state set, $S$). We define $\mathbf{f}_s^{(1)}$ as a $k_1$-dimensional binary vector, where $k_1$ represents the number of links in the road network, i.e., the number of states in $S$ ($k_1 = |S|$). A one-hot encoding is used to represent the $k_1$ different links in the network, where the $i^{th}$ element in $\mathbf{f}_s^{(1)}$ is set to 1 and all other elements are set to 0 to represent the $i^{th}$ link.
- The state feature vector, $\mathbf{f}_s^{(2)} \in \mathbb{R}^{k_2}$, is designed to convey information from the traffic volume data (i.e., the distribution of state visitation counts over the detector state subset, $S_v$). We define $\mathbf{f}_s^{(2)}$ as a $k_2$-dimensional binary vector, where $k_2$ represents the number of detector links in the road network, i.e., the number of states in $S_v$ ($k_2 = |S_v|$). For a detector link ($s \in S_v$), a one-hot encoding is used, where the $i^{th}$ element in $\mathbf{f}_s^{(2)}$ is set to 1 and all other elements are set to 0 to represent the $i^{th}$ detector link among the $k_2$ detectors. For a non-detector link ($s \in S\backslash S_v$), we simply assign a $k_2$-dimensional zero vector, where all elements in $\mathbf{f}_s^{(2)}$ are 0.

Based on the state feature vectors defined above, a state-action path $\tau$ is characterized by a path feature vector $\mathbf{f}_\tau \in \mathbb{R}^k$ ($k = k_1 + k_2$) by concatenating the following two path feature vectors:

$$\mathbf{f}_\tau = \left[\mathbf{f}_\tau^{(1)^T}, \mathbf{f}_\tau^{(2)^T}\right]^T \quad (7)$$

where $\mathbf{f}_\tau^{(1)} = \sum_{s\in\tau} \mathbf{f}_s^{(1)}$ becomes a $k_1$-dimensional binary vector indicating which links $\tau$ (or its corresponding vehicle trajectory) passes through and $\mathbf{f}_\tau^{(2)} = \sum_{s\in\tau} \mathbf{f}_s^{(2)}$ becomes a $k_2$-dimensional binary vector indicating which detectors $\tau$ passes through. For instance, consider a network with seven links ($k_1 = 7$), where three out of the seven links have detectors ($k_2 = 3$). If $\tau$ traverses the 1st, 4th, and 7th links, then $\mathbf{f}_\tau^{(1)}$ becomes $(1,0,0,1,0,0,1)$. Furthermore, if the 1st and 3rd detectors are these three links and, thus, $\tau$ passes through the 1st and 3rd detectors, then $\mathbf{f}_\tau^{(2)}$ becomes $(1,0,1)$. If calculate these path feature vectors for all state-action paths in some demonstrated path set $T_e$, then total path feature vector $\sum_{\tau\in T_e} \mathbf{f}_\tau^{(1)}$ tells us how many times each link is visited by the trajectories in $T_e$ (i.e., the $i$th element in $\sum_{\tau\in T_e} \mathbf{f}_\tau^{(1)}$ corresponds to the number of vehicle trajectories that pass the $i$th link). Similarly, total path feature vector $\sum_{\tau\in T_e} \mathbf{f}_\tau^{(2)}$ tells us how many times each detector is visited by the trajectories in $T_e$ (i.e., the $i$th element in $\sum_{\tau\in T_e} \mathbf{f}_\tau^{(2)}$ corresponds to the number of vehicle trajectories that pass the $i$th detector). It is important to note that the latter leads us to obtain the following relationship:

$$\sum_{\tau\in T_e} \mathbf{f}_\tau^{(2)} = \sum_{s\in S_v} v_s \mathbf{f}_s^{(2)} \quad (8)$$

where $v_s$ is the traffic volume on detector link $s$ produced by the trajectories in $T_e$, indicating that the sum of path feature vectors can be expressed in terms of link traffic volume observations. This relationship plays a key role in allowing traffic volume data to be used as expert demonstrations to guide the learning of a policy in the proposed IRL-F.

In the road-network MDP, the path reward is calculated the same way as in MaxEnt IRL (see Eq. (2)), where the state reward value that is linear to the state feature vector, parametrized by unknown reward weights $\theta \in \mathbb{R}^{k_1 + k_2}$. The path probabilities are defined the same way as in MaxEnt IRL (see Eq. (3)).

**IRL objective:** In the road-network MDP, let $T_P$ denote the ground-truth population trajectory set and $M$ denote the number of trajectories in $T_P$. To solve the link flow estimation problem, the goal is to find a policy (vehicles' link-to-link transition decisions) that produces the population trajectory distribution by finding the optimal reward weights $\theta^*$ that maximize the likelihood of population trajectories, as shown in Eq. (9). The gradient can be computed the same way as in MaxEnt IRL, as shown in Eq. (10), where the first part can be viewed as expert feature expectations and the second part can be viewed as policy feature expectation.

$$\theta^* = \underset{\theta}{\operatorname{argmax}}\, L = \underset{\theta}{\operatorname{argmax}} \sum_{\tau\in T_P} \log P(\tau) \quad (9)$$

$$\nabla_\theta L = \frac{1}{M}\sum_{\tau\in T_P} \mathbf{f}_\tau - \sum_{s\in S} D_s \mathbf{f}_s \\ = \mathbf{f}_{\text{expert}} - \mathbf{f}_{\text{policy}} \tag{10}$$

However, $T_P$ and $M$ are unknown to traffic modellers and, thus, we cannot directly calculate $\mathbf{f}_{\text{expert}}$. Instead, only the observed traffic volume data from a subset of links and the observed trajectory data from a subset of vehicles are available. Both types of observed data provide information about the population trajectory distribution to some extent. Based on our newly proposed state feature definitions, we express $\mathbf{f}_{\text{expert}}$ as the concatenation of two feature expectations, $\mathbf{f}_{\text{expert}}^{(1)}$ and $\mathbf{f}_{\text{expert}}^{(2)}$, to describe the expert's desired behaviour by leveraging the available trajectory and traffic volume data, respectively. The gradient can then be rewritten as follows:

$$\nabla_\theta L = \left[\frac{\sum_{\tau\in T_P} \mathbf{f}_\tau^{(1)^T}}{M}, \frac{\sum_{\tau\in T_P} \mathbf{f}_\tau^{(2)^T}}{M}\right]^T - \left[\sum_{s\in S} D_s \mathbf{f}_s^{(1)^T}, \sum_{s\in S} D_s \mathbf{f}_s^{(2)^T}\right]^T \\ = \left[\mathbf{f}_{\text{expert}}^{(1)\,T}, \mathbf{f}_{\text{expert}}^{(2)\,T}\right]^T - \left[\mathbf{f}_{\text{policy}}^{(1)\,T}, \mathbf{f}_{\text{policy}}^{(2)\,T}\right]^T \tag{11}$$

We propose to use the observed vehicle trajectory data to approximate $\mathbf{f}_{\text{expert}}^{(1)}$ and use the observed traffic volume data to approximate $\mathbf{f}_{\text{expert}}^{(2)}$, which will be discussed later. Similarly, the policy feature expectation $\mathbf{f}_{\text{policy}}$ is represented as the concatenation of $\mathbf{f}_{\text{policy}}^{(1)}$ and $\mathbf{f}_{\text{policy}}^{(2)}$. The goal of the proposed IRL-F is to match these two policy feature expectations with those two approximated expert feature expectations.

**Solution process:** The optimal reward weights ($\theta^*$) in IRL-F can be found using gradient-based optimization methods. The challenge is how to determine the gradient using only the observed traffic data and, specifically, how to approximate expert's feature expectations using the observed vehicle trajectories and traffic volume data.

Let $T_{obs}$ denote the set of observed vehicle trajectories. Since the population trajectory set is not available, we propose to approximate $\mathbf{f}_{\text{expert}}^{(1)}$ using the expected feature expectation over the observed trajectories in $T_{obs}$ as follows:

$$\mathbf{f}_{\text{expert}}^{(1)} = \frac{\sum_{\tau\in T_P} \mathbf{f}_\tau^{(1)}}{M} \approx \frac{\sum_{\tau\in T_{obs}} \mathbf{f}_\tau^{(1)}}{|T_{obs}|} \tag{12}$$

If the observed vehicle trajectories are representative of the population (i.e., the sampling rate is the same across the network and all paths with non-zero true flow has at least one observed trajectories), this approximation gives the true feature expectation $\mathbf{f}_{\text{expert}}^{(1)}$. Otherwise, they may deviate from the ground truth. However, as it is unlikely that the observed vehicle trajectory distribution will deviate dramatically from the population trajectory distribution, such approximated feature expectations can still guide the agent to find a realistic policy in the road-network MDP, which produces the state visitation count distribution that is close to the distribution from the population trajectory set.

The second feature expectation $\mathbf{f}_{\text{expert}}^{(2)}$ is designed to use the information from link traffic volume data. Based on our feature definition for $\mathbf{f}_s^{(2)}$ and Eq. (8), we obtain the relationship $\sum_{\tau \in T_P} \mathbf{f}_\tau^{(2)} = \sum_{s \in S_v} v_s \mathbf{f}_s^{(2)}$, where $v_s$ is the traffic volume observed on detector link $s$ under the population trajectories in $T_P$. We then propose to express $\mathbf{f}_{\text{expert}}^{(2)}$ using this relationship and further approximate $\mathbf{f}_{\text{expert}}^{(2)}$ by using an estimated number of the population trajectories ($\widehat{M}$) as follows:

$$\mathbf{f}_{\text{expert}}^{(2)} = \frac{\sum_{\tau \in T_P} \mathbf{f}_\tau^{(2)}}{M} = \frac{\sum_{s \in S_v} v_s \mathbf{f}_s^{(2)}}{M} \approx \frac{\sum_{s \in S_v} v_s \mathbf{f}_s^{(2)}}{\widehat{M}} \tag{13}$$

It is noted that the traffic volume data from loop detectors ($v_s$) capture all the vehicles passing the detectors (i.e., reflect the population trajectories) and, therefore, we only need to find a substitute for $M$ to obtain $\mathbf{f}_{\text{expert}}^{(2)}$.

Many previous studies have introduced different methods to infer population traffic flows from the observed data that capture a proportion of the population traffic. One of the most straightforward method is to find *scaling factors* to scale up the observed mobility data such as vehicle trajectories to match point traffic counts using certain optimization models (Iqbal et al. 2014, Miller et al, 2020). More complicated models involve machine learning methods and/or microsimulation (Huang et al., 2019, Alexander et al., 2015). In our study, to facilitate the implementation of the IRL-F model, the *controlled Least Absolute Deviation* (cLAD) method proposed by Miller et al. (2020) is chosen to find an estimated size of population trajectories ($\widehat{M}$) based on the observed data. This method is simple and computationally efficient because it does not involve any iterative procedures.

Let $P$ denote a set of origin-destination pairs indexed with $p$ and $S_v$ be a set of observed links indexed with $s$. In the cLAD method, two terms *local capture rate* and *system capture rate* are defined, where the local capture rate is the ratio of the observed number of trajectories $t_s$ to the observed traffic volume $v_s$ for each detector link $s \in S_v$ and the system capture rate, denoted by $r$, is the median of all local capture rates among the observed links (i.e., $r = \text{median}\left\{\frac{t_1}{v_1}, \frac{t_2}{v_2}, \dots, \frac{t_{|S_v|}}{v_{|S_v|}}\right\}$). The cLAD method determines the optimal scaling factors as follows:

$$\underset{x_p}{minimize} \sum_{s \in S_v} |e_s - v_s| + \gamma \sum_{p \in P} (x_p)^2 \tag{14}$$

$$s.t. \quad e_s = \sum_{p \in P} t_{s,p}\, \alpha_p \qquad \forall s \in S_v \tag{15}$$

$$\alpha_p = \frac{1}{r} + x_p \qquad \forall p \in P \tag{16}$$

where $e_s$ and $v_s$ are the estimated and observed traffic volume on detector link $s \in S_v$, respectively; $\alpha_p$ is the scaling factor for the observed trajectories between origin-destination pair $p \in P$; and $t_{s,p}$ is the number of observed trajectories between origin-destination pair $p$ while crossing detector link $s$. Constraint (16) shows that the scaling factor ($\alpha_p$) is determined based on the system capture rate $r$ and a free parameter $x_p$, which is used to penalize large deviations of such factor from $1/r$. Constraint (15) shows that the estimated traffic volume ($e_s$) is determined by multiplying the number of observed trajectories on detector link $s$ with the scaling factor. The objective function (14) then aims to find the optimal parameter $x_p$ that minimizes the difference between the traffic volume estimated based on the scaled trajectories ($e_s$) and the ground-truth traffic volume ($v_s$) over all detector links $s \in S_v$. The second term of Eq. (14) is added to minimize the statistical variance of scaling factors, where a hyperparameter $\gamma$ is introduced to control the statistical bias. Higher values of $\gamma$ can be used prevent overfitting.

The result from the optimization model above is the optimal parameter ($\tilde{x}_p$) that determines the trajectory scaling factors for each OD pair. Let $t_p$ denote the number of observed trajectories between OD pair $p$. The estimated number of the population trajectories ($\widehat{M}$) can then be obtained as follows:

$$\widehat{M} = \sum_{p \in P} \left( \frac{1}{r} + \tilde{x}_p \right) t_p \tag{17}$$

With this estimated $\widehat{M}$, the expert feature expectation $\mathbf{f}_{\text{expert}}^{(2)}$ can be approximated using Eq. (13).

The policy feature expectation $\mathbf{f}_{\text{policy}}$ can be obtained by concatenating $\mathbf{f}_{\text{policy}}^{(1)}$ and $\mathbf{f}_{\text{policy}}^{(2)}$ defined as follows:

$$\mathbf{f}_{\text{policy}}^{(1)} = \sum_{s \in S} D_s \mathbf{f}_s^{(1)} \tag{18}$$

$$\mathbf{f}_{\text{policy}}^{(2)} = \sum_{s \in S} D_s \mathbf{f}_s^{(2)} \tag{19}$$

where the state visitation frequency, $D_s$, for a given policy is estimated the same way as in MaxEnt IRL (see Algorithm 1 in Ziebart et al. (2008a)).

Finally, Table 1 shows the algorithm to solve IRL-F to find the optimal reward weights $\theta$ that produce the agent's policy feature expectation that best matches the expert's feature expectation using a gradient-based method, where the gradient is computed as the difference between the expert's and policy feature expectations obtained from Eqs. (11)-(19).

Table 1
Algorithm I: Solving IRL-F

**Input** the road-network MDP $M = (S, A, \mu_0, P_T, \gamma, H)$.
the set of state-action paths from the observed trajectory data $T_{obs}$;
the set of state visitation counts from the observed traffic volume data $Q_{obs}$;
number of iterations $T$; learning rate $\alpha$; convergence tolerance $\epsilon$
**Initialize** random reward weights, $\theta^1 \in \mathbb{R}^{k_1 + k_2}$
**Compute** expert feature expectation $\mathbf{f}_{\text{expert}} = \left[ \mathbf{f}_{\text{expert}}^{(1)\,T}, \mathbf{f}_{\text{expert}}^{(2)\,T} \right]^T$
**For** $t = 1$ **to** $T$:
Find a policy $\pi^t$ based on current reward weights $\theta^t$
Compute state visitation frequencies $D_s^t$ based on $\pi^t$
Compute policy feature expectation $\mathbf{f}_{\text{policy}}^t = \left[ \mathbf{f}_{\text{policy},}^{(1),t\;T}, \mathbf{f}_{\text{policy},}^{(2),t\;T} \right]^T$
Compute gradient $\nabla L(\theta^t)$
if $\nabla L(\theta^t) < \epsilon$ then
return $\theta^t$

end if
$\theta^{t+1} \leftarrow \theta^t + \alpha \nabla L(\theta^t)$
**End For**
**Return** $\theta^t$

**Link Flow Estimation:** The output of IRL-F is a reward function in the road-network MDP parameterized by the optimal reward weights. With this reward function, an optimal policy can be recovered using any RL method (e.g., dynamic programming method). Once the policy is found, state-action paths can be sampled on the road-network MDP. Each path generated by the agent can be viewed as a synthetic vehicle trajectory in the road network. A set of generated synthetic trajectories from this optimal policy reflect possible underlying population trajectories that produce the observed traffic patterns and, thus, can be used to obtain the traffic volume on each link to solve the link estimation problem in the road network.

With this ability to generate synthetic population trajectories, the goal of the proposed generative framework is to find the optimal set of synthetic vehicle trajectories that produce the best link flow estimates. The quality of a generated synthetic trajectory set can be evaluated by comparing the estimated link flows to the observed volume data for the detector links. To find the optimal size of the synthetic trajectory set, the most straightforward way is to keep generating trajectory sets with different scales until the one that minimises the difference between the estimated and observed volumes for the detector links is found. However, this method is computationally expensive due to a large number of candidate synthetic trajectory sets that need to be generated and evaluated.

In this paper, we propose an alternative method that utilises the state visitation frequencies, $D_s, \forall s \in S$, calculated during the process of solving IRL-F (see Algorithm I). Let $\widetilde{D}_s$ denote an estimate for $D_s$ obtained from the IRL-F model. Since $\widetilde{D}_s$ represents the probability of the agent visiting each state $s$ based on the policy learned by IRL-F, by assuming a uniform scaling factor $\beta \in \mathbb{R}$ across the states, we can translate $\widetilde{D}_s$ into the estimated traffic volume on each link $s$, denoted by $\tilde{v}_s$, through $\tilde{v}_s = \beta \widetilde{D}_s$. We use the following simple equation to calculate the optimal scaling factor, $\beta^*$:

$$\beta^* = \frac{1}{|S_v|} \sum_{s \in S_v} \frac{v_s}{\widetilde{D}_s} \tag{20}$$

which is the average scaling factor across the detector links obtained by first computing the ratio of the actual observed traffic volume ($v_s$) to the estimated state visit frequency ($\widetilde{D}_s$) for each detector link $s \in S_v$ and taking the mean over the detector link set ($S_v$). The traffic volume on an unobserved link ($\tilde{v}_s, \forall s \in S \backslash S_v$) can then be estimated by applying this scaling factor.

$$\tilde{v}_s = \beta^* \times \widetilde{D}_s, \qquad \forall s \in S \backslash S_v \tag{21}$$

### 3.3 Link flow estimation with CRL-F

Another approach we adopt to solve our road-network MDP is the Constrained Reinforcement Learning (CRL) method proposed by Miryoosefi et al. (2019). The CRL method allows the incorporation of a flexible constraint set to restrict the policies that the agent follows. In the road-network MDP, the information from the observed traffic data can be naturally expressed as constraints, with which a policy can be shaped in such a way to produce the state visitation frequencies that are consistent with the observed traffic volumes on the detector links. In this section, we will first briefly introduce the original CRL method and then describe our method (CRL-F) that applies the CRL to solve the link traffic estimation problem.

### 3.3.1 Constrained Reinforcement Learning

Given an MDP, the CRL defines the agent's learning goal in terms of *a vector of measurements* over the agent's behaviour, instead of a scalar *reward* function. When the agent chooses action $a_t$ at state $s_t$ and transitions to state $s_{t+1}$, it observes a *measurement vector*, $\mathbf{z}_t \in \mathbb{R}^d$, in a similar manner that the agent in the standard RL observes a reward, $r_t \in \mathbb{R}$. The measurement vector, $\mathbf{z}_t$, depends on the current state and action choice, i.e., $\mathbf{z}_t \sim P_z(\cdot\ |s_t, a_t)$. Actions are selected based on a policy $\pi$, which is the probability distribution over possible actions given state, $\pi(a_t|s_t)$. For any given policy $\pi$, the CRL defines the *long-term measurement vector*, $\bar{\mathbf{z}}(\pi) \triangleq \mathbb{E}[\sum_{t=0}^{\infty} \gamma^t\, \mathbf{z}_t|\pi]$, as the expected sum of discounted measurements for some discount factor $\gamma \in [0, 1)$. Considering a distribution of finitely many policies, which is referred to as a *mixed policy* $\mu$, the CRL further defines the *long-term measurement vector for mixed policy* $\mu$, $\bar{\mathbf{z}}(\mu) \triangleq \mathbb{E}_{\pi\sim\mu}[\bar{\mathbf{z}}(\pi)] = \sum_\pi \mu(\pi)\bar{\mathbf{z}}(\pi)$, as the expectation of $\bar{\mathbf{z}}(\pi)$ over different policies in $\mu$, where $\mu(\pi)$ is the probability of selecting policy $\pi$ from mixed policy $\mu$. Given the space of all single policies, $\pi \in \Pi$, and the space of all mixed policies $\mu \in \Delta(\Pi)$, the learning problem of the CRL is to find a feasible mixed policy $\mu$, whose long-term measurement vector $\bar{z}(\mu)$ lies in a convex constraint set $C \in \mathbb{R}^d$.

$$\text{Find } \mu \in\ \Delta(\Pi), \text{such that } \bar{\mathbf{z}}(\mu) \in C \tag{22}$$

In other words, the CRL can find a set of policies ($\mu$) that collectively satisfy a constraint that certain measurements produced by the agent ($\bar{\mathbf{z}}(\mu)$) become close to some target values ($C$).

The problem in Eq. (22) can be formulated as an optimization problem, with the objective to find a mixed policy $\mu \in\ \Delta(\Pi)$ so that the Euclidean distance between $\bar{\mathbf{z}}(\mu)$ and its closest point in the target set $C$ is minimized. This formulation can be expressed as follows:

$$\min_{\mu\in\ \Delta(\Pi)} \text{dist}(\bar{\mathbf{z}}(\mu), C) \tag{23}$$

where $\text{dist}(\cdot)$ is an operator that computes the Euclidean distance between a point and a set. The computation of this Euclidean distance can be further be written as a maximization problem. Assuming that the target set $C \in \mathbb{R}^d$ is a convex cone, the distance of any point $\mathbf{x} \in \mathbb{R}^d$ to this convex cone can be written as follows:

$$\text{dist}(\mathbf{x}, C) = \max_{\boldsymbol{\lambda}\in C^\circ \cap \mathcal{B}} \boldsymbol{\lambda} \cdot \mathbf{x} \tag{24}$$

where $C^\circ$ is the dual convex cone ($C^\circ \triangleq \{\boldsymbol{\lambda}: \boldsymbol{\lambda} \cdot \mathbf{x} \leq 0 \text{ for all } \mathbf{x} \in C\}$) and $\mathcal{B}$ is the Euclidean ball of radius 1 at the origin ($\mathcal{B} \triangleq \{\mathbf{x}: ||\mathbf{x}|| \leq 1\}$).

Given the expression of the Euclidean distance in Eq. (24), the CRL problem can be written in a min-max optimization form shown as follows,

$$\min_{\mu\in\ \Delta(\Pi)} \max_{\lambda\in\Lambda} \boldsymbol{\lambda} \cdot \bar{\mathbf{z}}(\mu) \tag{25}$$

where $\Lambda$ represents the space of $\boldsymbol{\lambda}$ defined as $\Lambda \triangleq C^\circ \cap \mathcal{B}$. This formulation can be interpreted as a special case of the two-person zero-sum game, where the $\mu$-player selects a mixed policy $\mu \in \Delta(\Pi)$ and the $\lambda$-player selects $\boldsymbol{\lambda} \in \Lambda$, resulting in a pay-out of $\boldsymbol{\lambda} \cdot \bar{\mathbf{z}}(\mu)$ from the $\mu$-player to the $\lambda$-player. The $\mu$-player wants to minimise this pay-out, while the $\lambda$-player wants to maximise it. Miryoosefi et al. (2019)

shows that, for the $\mu$-player to minimise $\boldsymbol{\lambda} \cdot \bar{\mathbf{z}}(\mu)$, it suffices to minimise $\boldsymbol{\lambda} \cdot \bar{\mathbf{z}}(\pi)$ over single policies $\pi \in \Pi$ and the problem of minimising $\boldsymbol{\lambda} \cdot \bar{\mathbf{z}}(\pi)$ is equivalent to finding a policy $\pi$ that maximises a long-term reward of $-\boldsymbol{\lambda} \cdot \bar{\mathbf{z}}(\pi)$ in a standard reinforcement learning task. The $\lambda$-player is modelled as a no-regret online learner who seeks to achieve small *regret*, the gap between its loss and the best in hindsight. Miryoosefi et al. (2019) proposed an algorithm called APPROPO (*approachability-based policy optimization*) to solve this two-person zero-sum game (see Algorithm 2 in the original paper). The outputs of this game are the average strategies selected by both players. Notably, the average strategy selected by the $\mu$-player (i.e., a uniform mixture of all single policies it selected at each iteration of this game) can be considered the optimal mixed policy for the CRL problem. An important aspect of this CRL method is that the definitions of both measurement vectors and the constraint set are flexible. The agent's behaviour can be trained towards different goals, which is achieved by defining different constraints. In Miryoosefi et al. (2019), the CRL was demonstrated in a grid-world environment, where three different measurement vectors were defined to restrict the agent's behaviour. Each of these measurement vectors corresponds to one constraint.

### 3.3.2 CRL-F

In this paper, we propose the CRL-F method, which applies the CRL approach to our problem, with the constraints specifically designed for the link flow estimation with trajectory and traffic volume data. The goal of the CRL-F is to find a mixed policy in the road-network MDP, based on which synthetic population vehicle trajectories are generated such that the state visitation frequencies produced by those trajectories conform the constraints imposed by the observed patterns in the available data.

In the CRL-F, two different measurement vectors are defined to characterise the constraints associated with the two types of data—trajectory data and traffic volume data—, respectively. The long-term measurement vector, $\bar{\mathbf{z}}(\pi)$, and the target set, $C$, are therefore considered to consist of two components as follows:

$$\bar{\mathbf{z}}(\pi) = \left[\bar{\mathbf{z}}^{(1)}(\pi)^T, \bar{\mathbf{z}}^{(2)}(\pi)^T\right]^T \text{ and } C = \left[C^{(1)^T}, C^{(2)^T}\right]^T \tag{26}$$

A single policy, $\pi \in \Pi$, induces a *long-term measurement vector* $\bar{\mathbf{z}}(\pi)$, which represents some collective information derived from the state-action paths generated by the agent following this policy. For instance, the state visitation frequency distribution produced by the generated state-action paths can be defined as a long-term measurement vector and, thus, in turn used as a constraint to control the agent's behaviour. By applying the same idea proposed for the IRL-F, we consider matching the state visitation frequency distributions between the agent (policy) and the data (expert) as our proposed constraints for the CRL-F.

The first long-term measurement vector, $\bar{\mathbf{z}}^{(1)}(\pi) \in \mathbb{R}^{k_1}$ corresponds to the first *policy* feature expectation, $\mathbf{f}_{\text{policy}}^{(1)}$, in the IRL-F.

$$\bar{\mathbf{z}}^{(1)}(\pi) \triangleq \mathbf{f}_{\text{policy}}^{(1)} = \sum_{s \in S} D_s \mathbf{f}_s^{(1)} \tag{27}$$

During the implementation of the APPROPO algorithm (Miryoosefi et al., 2019) to solve the CRL problem, a long-term measurement vector at each iteration can be estimated using the average of the measurement vectors collected over the previous iterations during the training. As such, we collect a set of state-action paths under policy $\pi$ generated during the training, denoted by $\hat{T}_\pi$, and take the average of the path feature vectors over this path set, $\hat{T}_\pi$, to estimate the state visitation frequencies ($D_s$, $\forall s \in S$) as follows:

$$D_s \approx \frac{\sum_{\tau \in \hat{T}_\pi} \sum_{s' \in \tau} \mathbf{1}_s(s')}{|\hat{T}_\pi|}, \qquad \forall s \in S \tag{28}$$

where the indicator function $\mathbf{1}_s(s')$ returns 1 if $s' = s$ and 0 if $s' \neq s$.

Following the original CRL method, we further define the *long-term measurement vector for mixed policy* $\mu$ as the expectation of $\bar{\mathbf{z}}^{(1)}(\pi)$ over different policies in $\mu$ as follows:

$$\bar{\mathbf{z}}^{(1)}(\mu) \triangleq \mathbb{E}_{\pi\sim\mu}[\bar{\mathbf{z}}^{(1)}(\pi)] \tag{29}$$

where mixed policy $\mu$ represents the distribution of single policies $\pi$. The target set associated with this first measurement vector, denoted by $C^{(1)}$, is then the state visitation frequency distribution obtained from the trajectory data, which corresponds to the first *expert* feature expectation, $\mathbf{f}_{\text{expert}}^{(1)}$, in the IRL-F, as follows:

$$C^{(1)} \triangleq \mathbf{f}_{\text{expert}}^{(1)} \approx \frac{\sum_{\tau \in T_{obs}} \mathbf{f}_\tau^{(1)}}{|T_{obs}|} \tag{30}$$

The first constraint of the CRL-F is to limit the Euclidean distance between the state visitation frequency distribution over all states in $S$ obtained from the model ($\mathbf{f}_{\text{policy}}^{(1)}$) and that obtained from the trajectory data ($\mathbf{f}_{\text{expert}}^{(1)}$) to be below a small threshold value denoted as $\epsilon_1$:

$$\text{Constraint I: } \text{dist}(\bar{\mathbf{z}}^{(1)}(\mu), C^{(1)}) = \text{dist}\left(\mathbf{f}_{\text{policy}}^{(1)}, \mathbf{f}_{\text{expert}}^{(1)}\right) < \epsilon_1 \tag{31}$$

Next, we define the second long-term measurement vector, $\bar{\mathbf{z}}^{(2)}(\pi) \in \mathbb{R}^{k_2}$, which corresponds to the second *policy* feature expectation, $\mathbf{f}_{\text{policy}}^{(2)}$, in the IRL-F. As with the first long-term measurement vector, $\bar{\mathbf{z}}^{(1)}(\pi)$, we use the generated state-action paths ($\hat{T}_\pi$) collected up to the current iteration during the training process of APPROPO to obtain an estimate for the state visitation frequencies ($D_s$, $\forall s \in S$).

$$\bar{\mathbf{z}}^{(2)}(\pi) \triangleq \mathbf{f}_{\text{policy}}^{(2)} = \sum_{s \in S} D_s \mathbf{f}_s^{(2)} \tag{32}$$

The corresponding *long-term measurement vector for mixed policy* $\mu$ is given by:

$$\bar{\mathbf{z}}^{(2)}(\mu) \triangleq \mathbb{E}_{\pi\sim\mu}[\bar{\mathbf{z}}^{(2)}(\pi)] \tag{33}$$

The target set associated with this second long-term measurement vector, denoted by $C^{(2)}$, is then defined as the state visitation frequency distribution obtained from the traffic volume data, which corresponds to the second *expert* feature expectation, $\mathbf{f}_{expert}^{(2)}$, in the IRL-F, as follows:

$$C^{(2)} \triangleq \mathbf{f}_{\text{expert}}^{(2)} \approx \frac{\sum_{s \in S_v} v_s \mathbf{f}_s^{(2)}}{\hat{M}} \tag{34}$$

The second constraint of the CRL-F limits the Euclidean distance between the state visitation frequency distribution over subset $S_v$ produced by the model ($\mathbf{f}_{\text{policy}}^{(2)}$) and that observed in the loop detector data ($\mathbf{f}_{\text{expert}}^{(2)}$) to be below another small threshold value denoted as $\epsilon_2$:

$$\text{Constraint II: } \text{dist}(\bar{\mathbf{z}}^{(2)}(\mu), C^{(2)}) = \text{dist}\left(\mathbf{f}_{\text{policy}}^{(2)}, \mathbf{f}_{\text{expert}}^{(2)}\right) < \epsilon_2 \tag{35}$$

Given the long-term measurement vectors and constraints described above, the objective of the CRL-F is to find a mixed policy $\mu$ that satisfies both Constraint I and Constraint II by following the same approaches used in the CRL as described in Eqs. (22)-(25), which can be solved using the APPROPO algorithm (Miryoosefi et al., 2019). Table 2 shows the algorithm implemented in this study, which was adapted from the original APPROPO algorithm. As mentioned above, APPROPO solves this problem as a two-person zero-sum game. At each iteration $t = 1, \cdots, T$, the $\mu$-player observes $\boldsymbol{\lambda}^t$ and finds a policy $\pi^t$ that maximises a long-term reward of $-\boldsymbol{\lambda}^t \cdot \bar{\mathbf{z}}(\pi^t)$ using standard reinforcement learning (RL). The resulting long-term measurement vectors for $\pi^t$ can be approximated, i.e., $\hat{\mathbf{z}}^t \approx \bar{\mathbf{z}}(\pi^t)$. An approach suggested by the authors of the APPROPO algorithm to speed up this procedure is to maintain a "cache" of policies, which can be used as "warm-starts" for solving RL or even eliminate the need for solving RL for every iteration. We adopt this approach and initialise a policy cache, a set of $N$ random policies and their associated long-term measurement vectors, at the beginning of the algorithm (line 3 of Table 2). Then, at each iteration $t = 1, \cdots, T$, we find the cached policy that has the maximum reward given current $\boldsymbol{\lambda}^t$ (line 5). If the selected cached policy is already good enough, there is no need to generate a new policy (lines 6-8). If not, we find a new policy by running a *Deep Q-learning algorithm*, an RL algorithm that uses a *neural network* to approximate *Q-function*, which is a function that returns the overall expected reward given a certain state and action pair, required to find the optimal policy in RL (line 10). More specifically, each policy is specified as a Q-function, which is approximated by a fully-connected neural network with three layers: input-layer representing states, one hidden layer, and output layer representing actions. The output of the deep Q-learning algorithm is an approximately optimal policy ($\pi^t$) that maximises a long-term reward of $-\boldsymbol{\lambda}^t \cdot \bar{\mathbf{z}}(\pi^t)$. Once the long-term measurement vectors are obtained using $\hat{\mathbf{z}}^t \approx \bar{\mathbf{z}}(\pi^t)$ (line 11), the $\lambda$-player, which is a no-regret online learner, then incurs loss $L_t(\boldsymbol{\lambda}^t) = -\boldsymbol{\lambda}^t \cdot \hat{\mathbf{z}}^t$ and updates $\boldsymbol{\lambda}^t$ to achieve small *regret*, which is defined as $\text{Regret}^T \triangleq [\sum_{t=1}^{T} L_t(\boldsymbol{\lambda}^t)] - \min_{\boldsymbol{\lambda} \in \Lambda} [\sum_{t=1}^{T} L_t(\boldsymbol{\lambda})]$, by playing a no-regret online learning algorithm such as *online gradient descent* (line 13). The output of the APPROPO algorithm is an estimated mixed policy, $\bar{\mu}$, which is a uniform mixture over all policies obtained during the T iterations $(\pi^1, \cdots, \pi^T)$.

Table 2

Algorithm II: Solving CRL-F using APPROPO algorithm (adapted from Miryoosefi et al., 2019)

1: **Input** projection $\Gamma_\Lambda(\mathbf{x}) = \text{argmin}_{\mathbf{x}' \in \Lambda} \|\mathbf{x} - \mathbf{x}'\|$, number of iterations $T$, step size $\eta$, cache size $N$, reward tolerance $\epsilon_0$, estimation tolerance $\epsilon_1$

2: **Initialize** $\boldsymbol{\lambda}^1$ arbitrarily in $\Lambda$

3: **Initialize policy cache** $\boldsymbol{c}$ by building a cache with $N$ random policies together with approximated long-term measurement vector

$$\boldsymbol{c} = \{\{\pi_{\boldsymbol{c}}^1, \bar{\mathbf{z}}(\pi_{\boldsymbol{c}}^1)\}, \{\pi_{\boldsymbol{c}}^2, \bar{\mathbf{z}}(\pi_{\boldsymbol{c}}^2)\}, \dots, \{\pi_{\boldsymbol{c}}^N, \bar{\mathbf{z}}(\pi_{\boldsymbol{c}}^N)\}\}$$

4: **For** $t = 1$ **to** $T$:

5: Find $\pi_{\boldsymbol{c}}^k$ in the policy cache $\boldsymbol{c}$, where $k = \text{argmax}_{i=1,2,\dots,N} \left(-\boldsymbol{\lambda}^t \cdot \bar{\mathbf{z}}(\pi_{\boldsymbol{c}}^i)\right)$

6: **If** $-\boldsymbol{\lambda}^t \cdot \bar{\mathbf{z}}(\pi_{\boldsymbol{c}}^k) \geq -\epsilon_0$ **then**

7: Set $\pi^t = \pi_{\boldsymbol{c}}^k$;

8: Set $\hat{\mathbf{z}}^t \approx \bar{\mathbf{z}}(\pi^t)$

9: **Else then**

10: Using the $\pi_{\boldsymbol{c}}^k$ as a warm-start, find a policy $\pi^t$ that satisfies $-\boldsymbol{\lambda}^t \cdot \bar{\mathbf{z}}(\pi^t) \geq -\epsilon_0$ by applying a deep Q-learning algorithm, compute the approximated long-term measurement vector $\bar{\mathbf{z}}(\pi)$ for each policy $\pi$

11: Set $\hat{\mathbf{z}}^t \approx \bar{\mathbf{z}}(\pi^t)$

12: Update the policy cache by adding the newly generated policy and its approximated measurement vectors $\{\pi^t, \bar{\mathbf{z}}(\pi^t)\}$.

13: Update $\boldsymbol{\lambda}^t$ using online gradient descent with the loss function $L_t(\boldsymbol{\lambda}) = -\boldsymbol{\lambda} \cdot \hat{\mathbf{z}}^t$:
$\boldsymbol{\lambda}^{t+1} \leftarrow \Gamma_\Lambda(\boldsymbol{\lambda}^t - \eta \nabla L_t(\boldsymbol{\lambda}^t)) = \Gamma_\Lambda(\boldsymbol{\lambda}^t + \eta \hat{\mathbf{z}}^t)$

14: **If** $L_t(\boldsymbol{\lambda}^t) < -(\epsilon_0 + \epsilon_1)$ **then**

15: **Return** problem is not feasible

16: **End If**

17: **End For**

18: **Return** $\bar{\mu}$, a uniform mixture over $\pi^1, \cdots, \pi^T$

**Link Flow Estimation:** The output of the CRL-F is the optimal mixed policy, $\bar{\mu}$, which is a uniform mixture of single policies generated across all iterations until the algorithm terminates. One the mixed policy is found, synthetic vehicle trajectories can be generated by first sampling a single policy randomly from this mixed policy and then by letting the agent in the road-network MDP make sequential decisions based on this single policy starting from an initial state sampled from the initial state distribution. The synthetic vehicle trajectories generated from the mixed policy collectively satisfy the two constraints associated with the trajectory and traffic volume data, reflecting possible underlying population trajectories that produce the observed traffic patterns.

With this ability to generate synthetic population trajectories, the next step is to find the optimal set of synthetic trajectories that lead to the best link flow estimates, as in the IRL-F. We can apply the same approach used in the IRL-F, which identifies the scaling factor ($\beta^*$) that maps the state visitation frequencies ($\widetilde{D}_s$) obtained from synthetic trajectories to the actual population link volumes ($\tilde{v}_s$) by using Eqs. (20) and (21). In the CLR-F, the estimated state visitation frequencies, $\widetilde{D}_s$, can be computed based on a set of trajectories $T_{\bar{\mu}}$ generated under mixed policy $\bar{\mu}$ by enumerating all paths in $T_{\bar{\mu}}$ and counting the number of times each state $s$ is visited as follows:

$$\widetilde{D}_s = \frac{D_s^{T_{\bar{\mu}}}}{\sum_{s \in S} D_s^{T_{\bar{\mu}}}}, \quad \forall s \in S \tag{36}$$

where $D_s^{T_{\bar{\mu}}} = \sum_{\zeta \in T_{\bar{\mu}}} \sum_{s' \in \zeta} \mathbf{1}_s(s')$.

## 4. Experiments

The proposed generative modelling framework has been applied to solve the link flow estimation problem in different test networks. First, the proposed framework is validated by estimating link flows on a real road network, namely the Berlin-Friedrichshain network. Next, a comparative analysis is performed by comparing the performance of our framework with that of two existing models in the literature using the Nguyen-Dupuis network.

### 4.1 Model Validation

**Study network and data:** To validate the proposed framework, we used the Berlin-Friedrichshain network, which is a road network data in Berlin available in a public repository (Transportation Networks for Research Core Team). The road network is shown in Figure 3, which include 224 nodes and 523 links. The dataset includes OD demand as well as the parameters of link performance functions. To obtain ground-truth link flows and trajectory data, we conducted traffic assignment using the given

OD demand and link performance functions, assuming User Equilibrium (UE) conditions. The resulting ground-truth link flows are shown in Figure 4, where the width of each link is weighted by its traffic volume. The traffic assignment results produced population trajectories covering a total of 1616 paths.

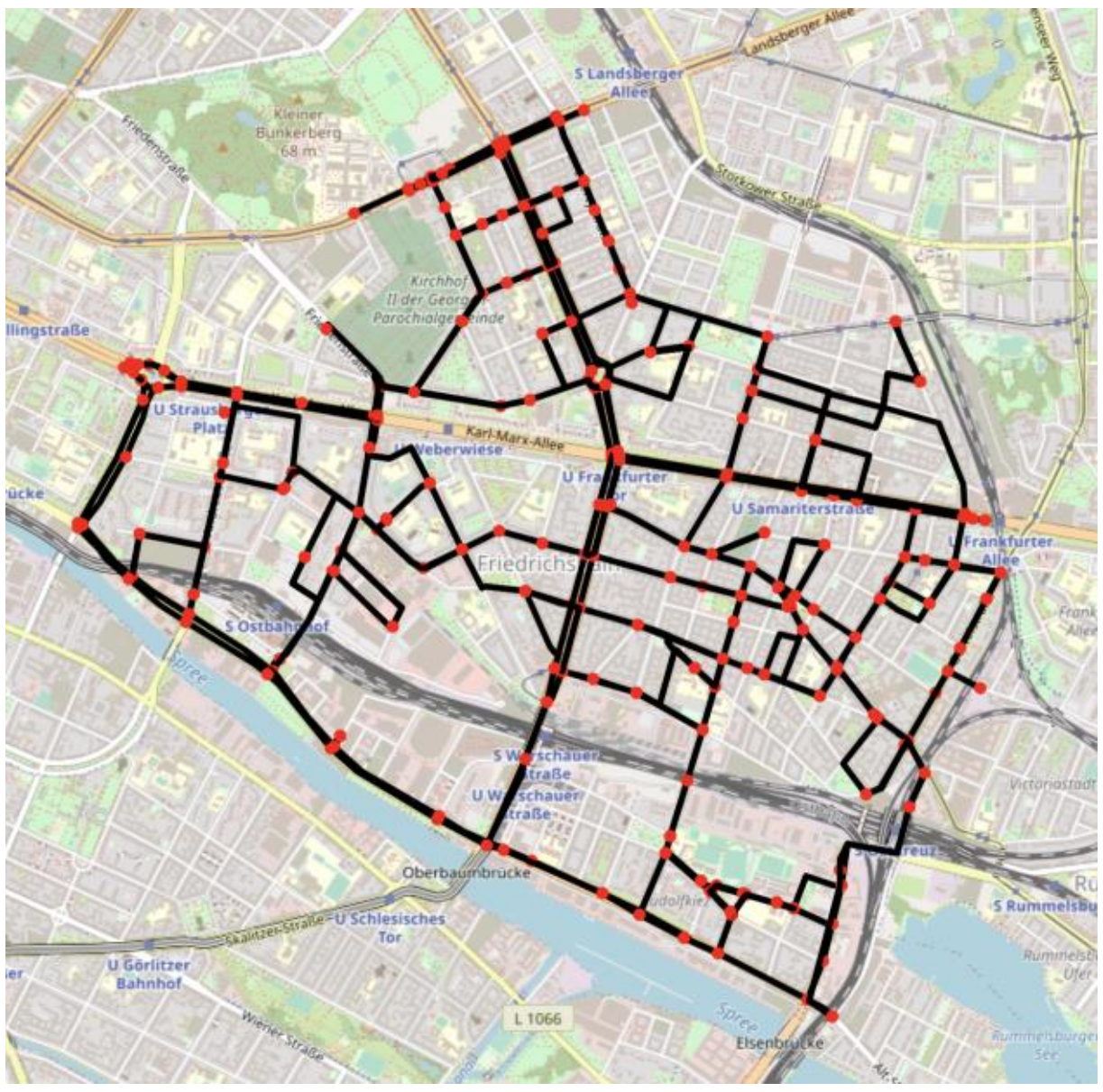

Figure 3. The Berlin-Friedrichshain network

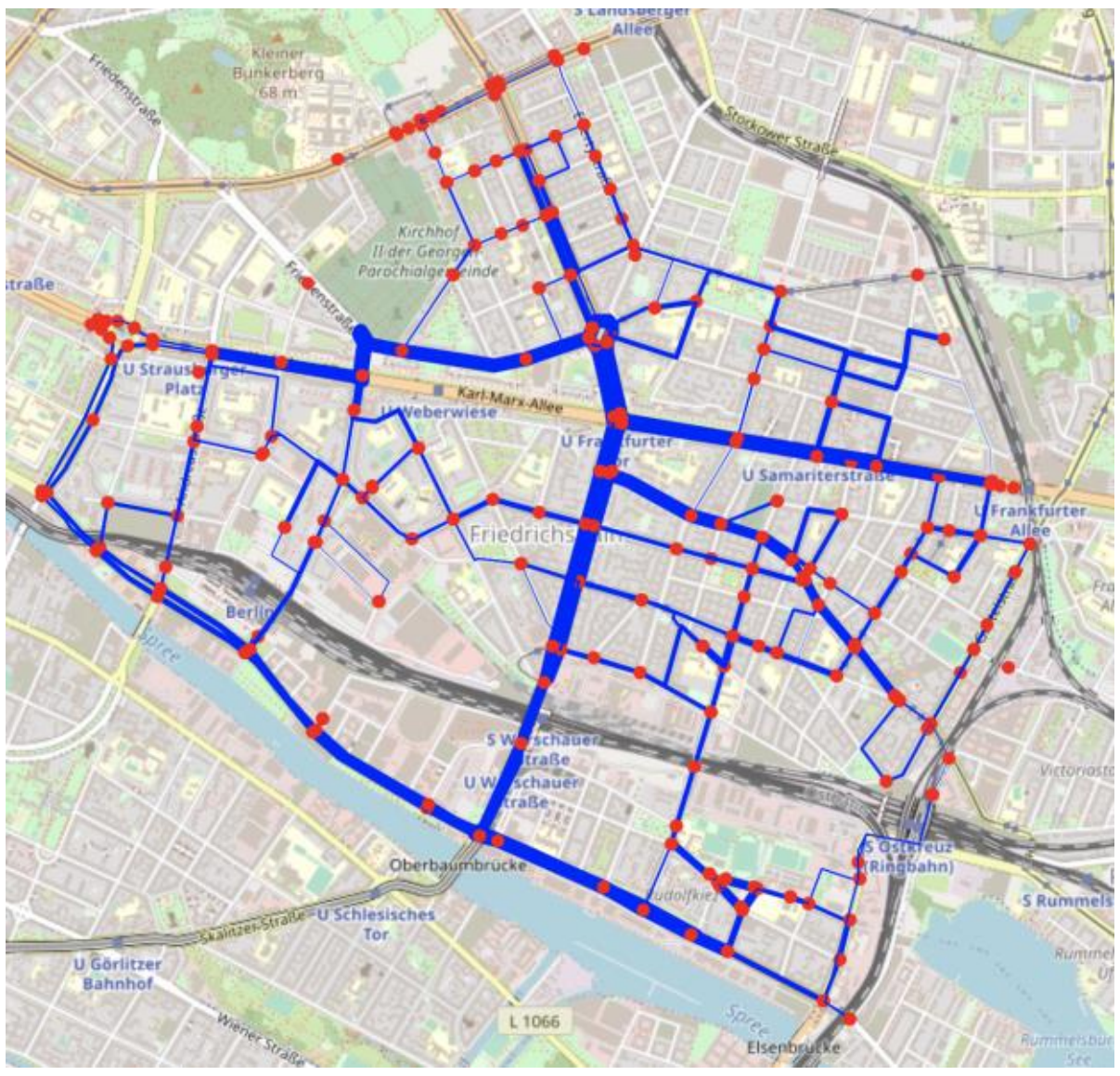

Figure 4. Ground-truth traffic flow in the Berlin-Friedrichshain network

**Traffic volume data input:** Once the ground-truth link flows and population trajectories (path flows) are obtained, the set of links in the network is divided into two subsets: *observed links* ($S_v$) and *unobserved links* ($S \backslash S_v$). For the observed links, which represent links with loop detectors, we randomly select 30% of the links (157 links) and the link flow data from those 157 links are used as the traffic volume data input to our framework.

**Trajectory data input and sampling rates:** To prepare the trajectory data input, firstly a range of sampling rates is set to consider in the validation study. The sampling rate for each of the 1616 paths is determined by randomly drawing a sampling rate from the given range. The number of observed trajectories for each path is then calculated by multiplying the original path flow by the sampling rate for that path. In order to create test scenarios where some paths in the road network are not covered by observed trajectories, we randomly select 5% of the paths (81 paths) with non-zero true path flows to have zero observed trajectory. Additionally, some paths may have no observed trajectories if the computed number of trajectories after applying the sampling rate is less than 1. The observed trajectories sampled this way form the sample trajectory data input to our framework. In this study, we consider two sampling rate ranges: the range of 20 - 40 % and the range of 10 - 30 %, which are selected to reflect sparse trajectory datasets in real-world situations.

**Feature definition:** As described in the IRL-F section, we use unique link IDs to define two state feature vectors, $\mathbf{f}_s^{(1)}$ and $\mathbf{f}_s^{(2)}$, associated with trajectory data and traffic volume data, respectively. In the original MaxEnt IRL, however, state feature vector is defined in terms of general characteristics describing each road segment. To evaluate the usefulness of the proposed (unique ID-based) feature definition over such a conventional feature definition, we test a scenario where $\mathbf{f}_s^{(1)}$ is defined in terms of general road characteristics such as *road type*, *road length*, and *maximum travel speed* for comparison. It is noted that $\mathbf{f}_s^{(2)}$ is defined using unique link IDs for all test scenarios because $\mathbf{f}_s^{(2)}$ is specifically designed to capture traffic volume data from detector links, which is unique to our link flow estimation problem. Table 3 shows the three features and their definitions considered in this study. All features are represented as categorical variables and the categories of each variable are defined based on the road segment data for the Berlin-Friedrichshain network obtained from the OpenStreetMap (OSM), which is an open-source map available online (OpenStreetMap contributors). To construct $\mathbf{f}_s^{(1)}$, we express each feature as a binary vector using a one-hot encoding and concatenate the three binary vectors from the three features to form one state feature vector. For instance, $\mathbf{f}_s^{(1)}$ for a link with a road type of freeway, a road length of 300 m, and a maximum travel speed of 60 km/h is specified as:

$$\mathbf{f}_s^{(1)} = [\underbrace{1,0,0,0}_{\text{type}}, \underbrace{0,1,0,0}_{\text{length}}, \underbrace{0,0,1,0}_{\text{max. speed}}]$$

Table 3. Features describing general road characteristics

| Feature | Value |
|---|---|
| Road type | {freeway, arterial road, collector, local road} |
| Road length (m) | {0-100, 100-500, 500-1000, >1000} |
| Maximum travel speed (km/h) | {40, 50, 60, >60} |

**Test Scenarios:** The goal of this validation study is to evaluate how closely the estimated link flows on the unobserved links agree with the ground-truth link flows. We consider the following four scenarios under different trajectory sampling rates and state feature definitions to validate the proposed generative modelling framework, which is based on either IRL-F or CRL-F:

- Scenario-A1: the trajectory sampling rates range between 20% and 40%; state feature vector $\mathbf{f}_s^{(1)}$ is defined using unique link IDs (proposed method).
- Scenario-A2: the trajectory sampling rates range between 20% and 40%; state feature vector $\mathbf{f}_s^{(1)}$ is defined using the general road characteristics in Table 3.
- Scenario-A3: the trajectory sampling rates range between 10% and 30%; state feature vector $\mathbf{f}_s^{(1)}$ is defined using unique link IDs (proposed method).
- Scenario-A4: the trajectory sampling rates range between 10% and 30%; state feature vector $\mathbf{f}_s^{(1)}$ is defined using the general road characteristics in Table 3.

For all test scenarios, the generative models using IRL-F were trained in a cluster node using 4 Intel cores and 20 GB of memory, and the generative models using CRL-F were trained in a cluster node using Intel Xeon Gold 6132, 140 GB of memory, and NVIDIA Tesla V100.

The performance of the proposed model under each test scenario is measured using the weighted absolute percentage error (WAPE). For the set of unobserved links, denoted by $S_u$, i.e., $S_u = \{s|s \in S \backslash S_v\}$, the value of WAPE can be calculated as follows:

$$WAPE = \frac{\sum_{s \in S_u} |\tilde{v}_s - v_s|}{\sum_{s \in S_u} v_s} \quad (37)$$

where $\tilde{v}_s$ is the estimated link flow on link $s$ and $v_s$ is its ground-truth link flow.

**Results:** We first show the graphical comparison of the estimated link flows and ground-truth link flows for the tested scenarios in Figure 5, where the x-axis represents the IDs of the unobserved links sorted in ascending order by the ground-truth link flow values. The major finding based on the visual inspection is that the use of the unique link ID features (Scenarios A1 & A3) produces a much better agreement between the estimated and ground-truth link flows than the use of the general road characteristic features (Scenarios A2 & A4). This demonstrates the effectiveness of our proposed feature definition method in the context of the link flow estimation problem as it allows the feature expectation matching, which is the mechanism used in both IRL-F and CRL-F, to directly learn the link flow patterns in trajectory data. Next, the performance comparison based on WAPE is shown in Table 4. The performance difference under different feature definitions can be further confirmed, where the WAPE of both IRL-F and CRL-F models are much smaller in Scenarios A1 & A3 when compared to the results in Scenarios A2 & A4. In terms of trajectory sampling rate, we observe that the decrease in the sampling rate from [20%, 40%) to [10%, 30%) tends to decrease the estimation accuracy (e.g., results in WAPE are overall higher in Scenario A3 than in Scenario A1 and, similarly, in Scenario-A4 than in Scenario-A2). This indicates that the lower the sampling rate of the available trajectory data is, the less likely the data are to be representative of the population and, thus, the more challenging it is to recover the population link flow patterns from data. Despite the low sampling rates, however, the proposed IRL-F and CLR-F models with the unique link feature definition (Scenarios A1 & A3) generate the link flow estimates that are well in agreement with the ground-truth data as shown in Figure 5. Between the IRL-F and CRL-F, there is no significant difference in the model performance, indicating that both IRL-F and CRL-F can serve as a synthetic trajectory generator within the proposed generative modelling framework for estimating link flows in road networks.

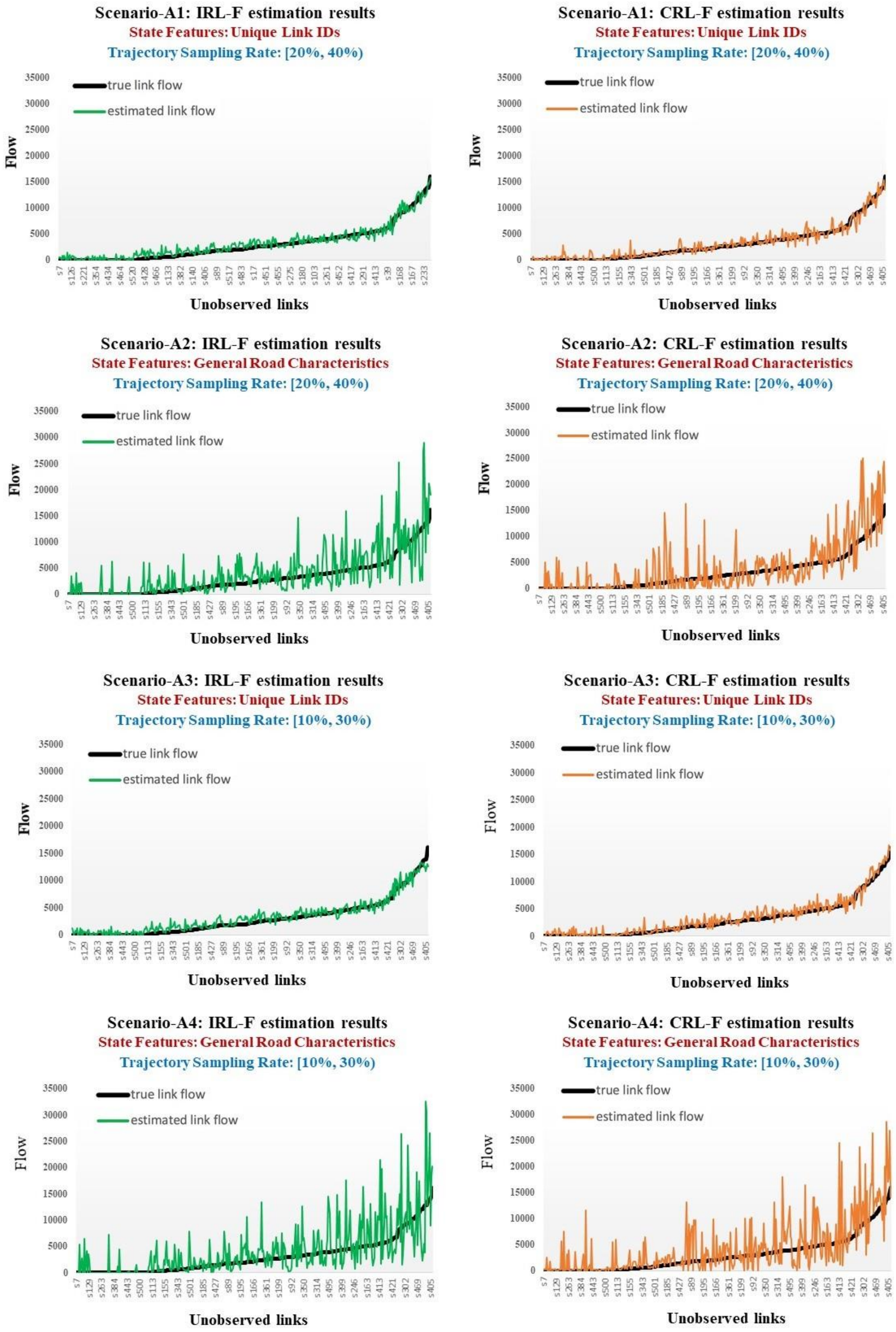

Figure 5. Link flow estimation results in the Berlin-Friedrichshain network

Table 4. Performance of the proposed model in test scenario set A

| Scenario | Model specification | WAPE |
|---|---|---|
| Scenario-A1<br>State Features: Unique Link IDs<br>Trajectory Sampling Rate: [20%, 40%) | IRL-F | 19.75% |
| | CRL-F | 17.55% |
| Scenario-A2<br>State Features: General Road Characteristics<br>Trajectory Sampling Rate: [20%, 40%) | IRL-F | 65.70% |
| | CRL-F | 66.55% |
| Scenario-A3<br>State Features: Unique Link IDs<br>Trajectory Sampling Rate: [10%, 30%) | IRL-F | 21.32% |
| | CRL-F | 19.37% |
| Scenario-A4<br>State Features: General Road Characteristics<br>Trajectory Sampling Rate: [10%, 30%) | IRL-F | 73.22% |
| | CRL-F | 76.83% |

### 4.2 Model Comparison

**Benchmark models.** In this section, we compare the proposed generative modelling framework (solved by either IRL-F or CRL-F) to two of the existing methods in the literature as benchmarks.

- The first method is proposed by Zhou and Mahmassani (2006), referred to as *ZM* hereafter, which solves the OD demand estimation problem using traffic volume data and Automatic Vehicle Identification (AVI) data, where AVI data can be viewed as a special type of vehicle trajectory data. The link flow estimation is a by-product of this OD estimation process since traffic assignment is conducted at each step in the proposed iterative solution algorithm. The observed traffic data are used in a multi-objective optimization framework, which requires prior knowledge of OD demands as inputs. Note that the solution algorithm proposed in this paper requires a traffic simulator to solve the traffic assignment problem.
- The second method is proposed by Brunauer et al. (2017), referred to as *BHR* hereafter, which solves link flow estimation using traffic volume data and probe vehicle trajectories. This method considers the link flow estimation problem as a local network propagation problem. The unobserved link flows are estimated by propagating observed link flows across the network based on propagation rules defined using the observed trajectories. Note that, to calculate the propagation rules across the network, this method assumes that the observed trajectory data cover most of the link-to-link transitions in the road network.

**Study network and parameter settings:** The comparison analysis is conducted using the Nguyen-Dupuis network, which consists of 13 nodes and 38 links, as shown in Figure 6. We chose the Nguyen-Dupuis network for the comparison analysis, instead of the Berlin-Friedrichshain network, because the coverage of the ground-truth path flows in the Berlin-Friedrichshain network is relatively small thereby making it not suitable for implementing the BHR model, which requires trajectory data with high spatial coverage. For the Nguyen-Dupuis network, we used the network descriptions and OD and flow data provided by Castillo et al. (2008). Among 38 links, 11 links are selected as observed links with detectors ($S_v$) as shown in Figure 6, which is 29% of the links in the network. The ground-truth link flows from these observed links are used as traffic volume data input to the link estimation problem. For the trajectory data input, we apply the same sampling procedure used in the validation study above to sample observed trajectories from the ground-truth path flows. Among all paths with non-zero ground-truth path flows, we randomly select 5% of the paths and assign them zero observed trajectory to create situations where not all paths are covered by the observed trajectory data. For the rest of the paths, we consider the following nine sampling rate intervals: [5%, 15%), [5%, 25%), [5%, 35%), [15%, 25%), [15%, 35%), [15%, 45%), [25%, 35%), [25%, 45%), [25%, 55%). For the feature definition for the IRL-F and CRL-F, we use our proposed method that uses the unique link IDs.

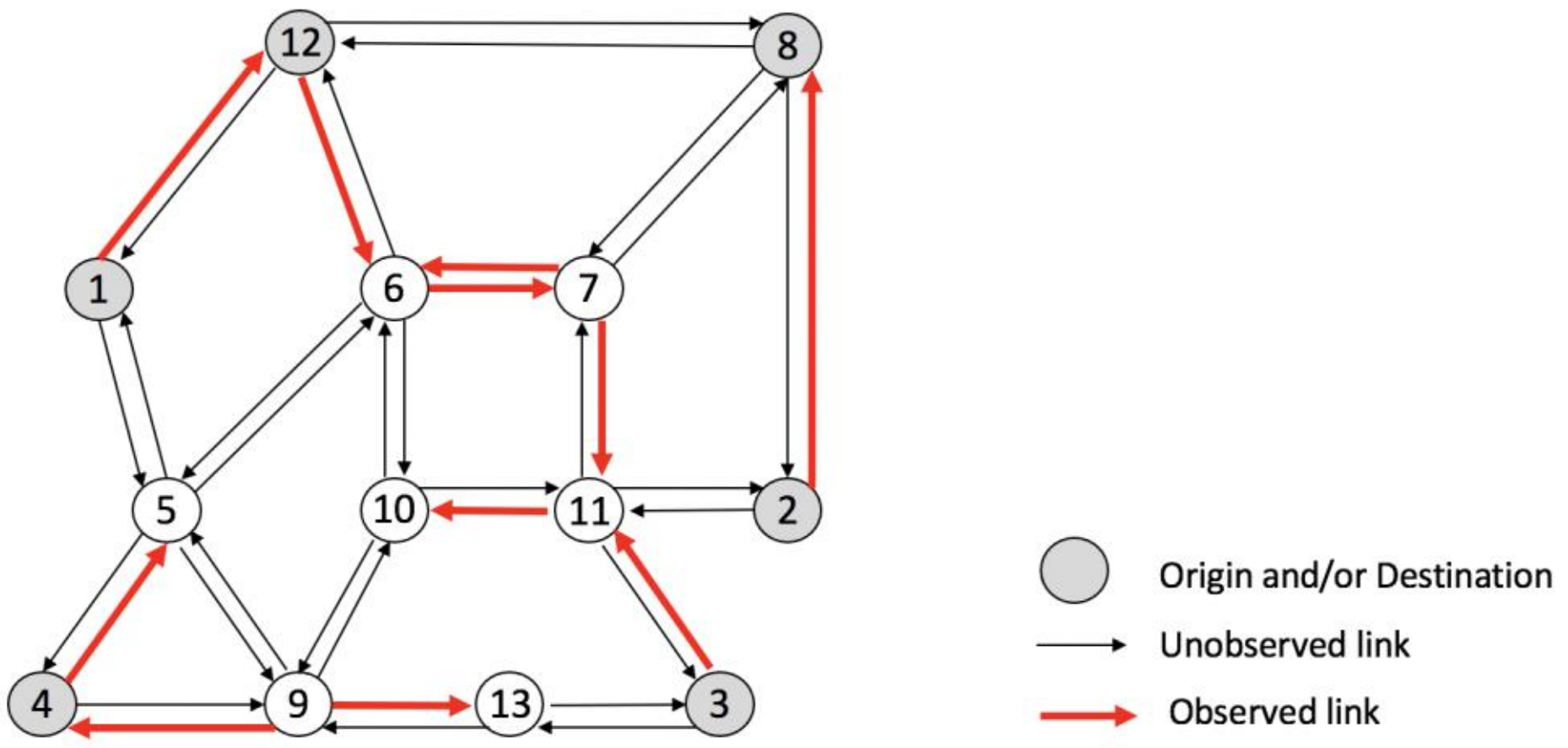

Figure 6. The Nguyen-Dupuis network

**Test scenarios and performance measure:** Each of the two benchmark models, ZM and BHR, requires some specific assumptions to be met. For example, the ZM model assumes that the traffic simulator successfully captures the true traffic flow patterns so the observed traffic data (e.g., link count data and AVI data) are not too much deviated from the estimated traffic flows generated by the simulator. The BHR model assumes that the observed trajectories cover most of the link-to-link transitions in the study network. To allow fair comparisons, we design two sets of scenarios for each benchmark model: the first scenario set provides the conditions required by a given benchmark model, whereas the second scenario set lifts these requirements to test the model performances in a more flexible and realistic environment. Each set is further divided into nine individual scenarios to test the nine different trajectory sampling intervals mentioned above. For all test scenarios, the generative models using either IRL-F or CRL-F were trained in the same environment as in the validation experiments discussed in Section 4.1.

### 4.2.1 Comparison with the ZM (Zhou-Mahmassani) model

Two sets of test scenarios are designed to compare our proposed model to the ZM model: Scenario set B, which assumes the traffic simulator used in the ZM model captures the true flow patterns, and scenario set C, which relaxes this assumption.

Note that, besides the traffic volume data and the vehicle trajectory data, the ZM model further requires prior knowledge of the OD demand. To investigate the performance of the models with respect to different levels of prior knowledge, three different prior OD demand settings are designed as follows:

- *P1*: the prior demand for each OD pair is equal to its true demand value with small disturbances by adding an error sampled from a normal distribution with zero mean and a standard deviation of 20% of the true demand value.

- *P2*: the prior demand for each OD pair is equal to its true demand value with moderate disturbances by adding an error sampled from a normal distribution zero mean and a standard deviation of 40% of the true demand value.
- *P3*: the same prior demand is used for all OD pairs, where the common prior demand value is obtained by taking the average of the true demands over all OD pairs.

Among the three settings, *P1* reflects the most accurate prior knowledge of the OD demand, whereas *P3* reflects the least accurate prior knowledge.

**Scenario set B:** It is assumed that the traffic simulator used to implement the ZM model conducts traffic assignment under static user equilibrium conditions. Meanwhile, the ground-truth traffic observations on the Nguyen-Dupuis network are obtained by conducting traffic assignment under the same user equilibrium conditions. By doing this, the traffic simulator can be considered to capture the true traffic flow pattern as the simulated traffic flow patterns would be consistent with the ground-truth traffic flow patterns. Figure 7 shows the comparison results using the WAPE between the proposed generative modelling framework (solved using either IRL-F or CRL-F) and the ZM model tested under the nine trajectory sampling rates. For each trajectory sampling rate, the ZM model is conducted under the aforementioned three prior OD demand settings (P1-P3). As shown in Figure 7, IRL-F and CRL-F achieve similar performance within each test scenario. When comparing across different scenarios, the estimation errors are smaller when trajectory sampling rates are higher (e.g., the errors are smaller in Scenario B7 than in Scenario B1) or have a smaller range (e.g., the errors are smaller in Scenario B7 than in Scenario B9), meaning that the observed data provides better distribution information about the population trajectories when sampling rates are higher and less heterogeneous. The ZM model shows a similar level of errors across different trajectory sampling rates, indicating that it is less sensitive to the trajectory sampling rate. This is largely because the multi-objective optimisation technique used in the ZM model, which adjusts the objective function weights associated with different input sources, give less weight to the trajectory data while giving more weight to other data sources such as link counts and prior OD demand in this particular experiment. Due to its high reliance on the prior OD demand, the performance varies with the prior OD demand setting; the estimation error increases as the mismatch between the prior information on the OD demand and the true demand increases (from P1 to P3). Our models (IRL-F and CRL-F) outperform the ZM model when the trajectory sampling rates are higher and have a narrower range (e.g., Scenarios B4, B7 and B8). The ZM model shows the similar or better performance in some cases, especially when the prior OD is closer to the truth demand (e.g., in P1). Overall, our models and the ZM model show comparable performance in scenario set B, where the assumption for the traffic simulator required by the ZM method is satisfied.

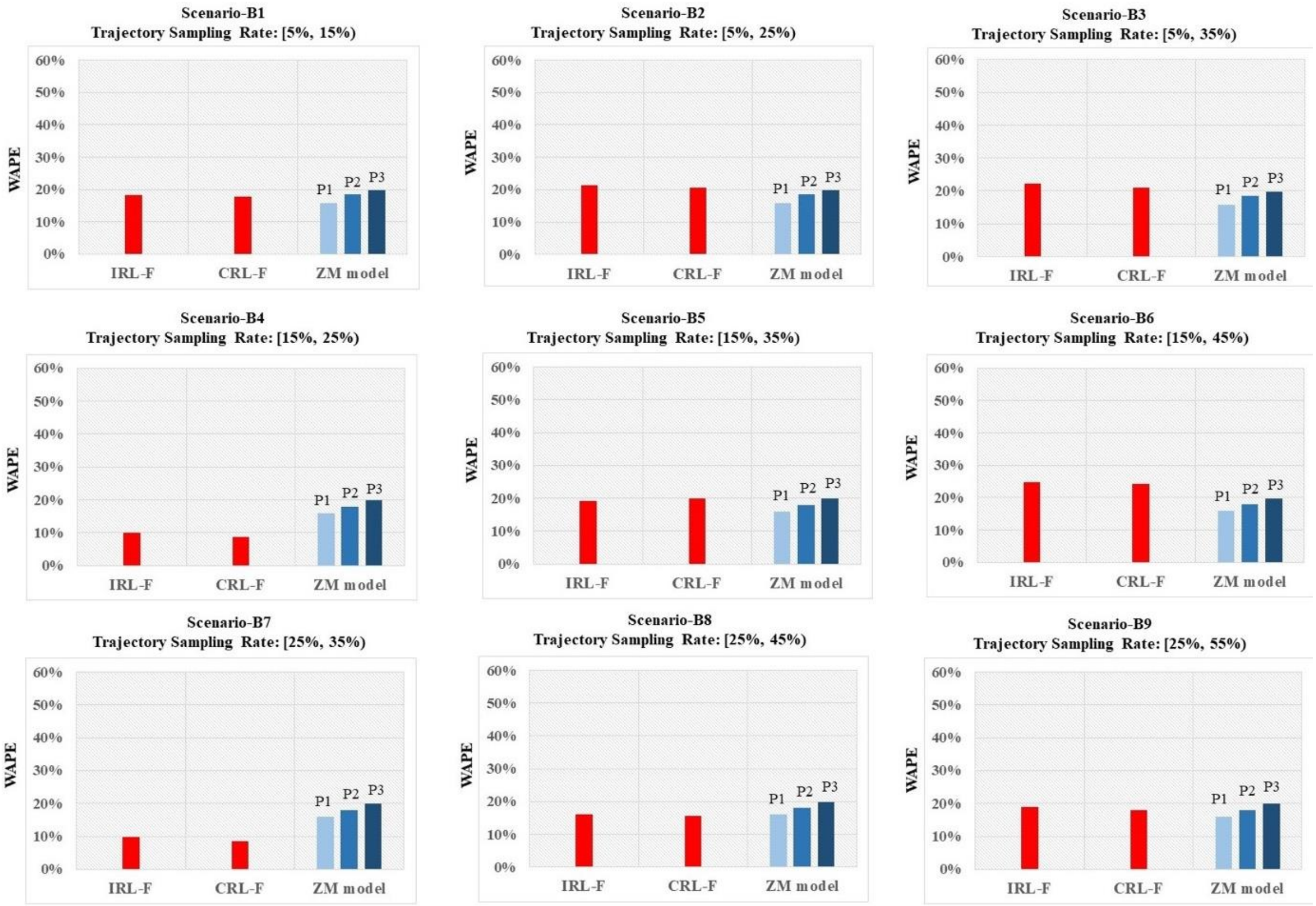

Figure 7. Performance comparison with ZM model for scenario set B

**Scenario set C:** In this scenario set, the assumption that a traffic simulator can capture the true traffic flow pattern is relaxed. To create this test environment, we use two different path cost functions to perform traffic assignment in generating the ground-truth traffic flow and in implementing the ZM model, respectively, so that the simulator used in implementing the ZM model cannot accurately replicate the ground-truth traffic flow pattern. Specifically, we assume that the true traffic flows on the Nguyen-Dupuis network are from the traffic assignment results based on a path cost function $C_p^1$, while a simulator used to implement the ZM model performs traffic assignment based on another slightly different path cost function $C_p^2$. These cost functions are defined as follows:

$$C_p^1 = \sum\nolimits_{l \in L} \delta_l^p C_l(f_l) + T_p\left(\sum\nolimits_{l \in L} \delta_l^p w_l\right) \qquad \forall p \in P \tag{38}$$

$$C_p^2 = \sum\nolimits_{l \in L} \delta_l^p C_l(f_l) + \mu \sum\nolimits_{l \in L} \delta_l^p w_l \qquad \forall p \in P \tag{39}$$

where $P$ is the set of all possible paths in the test network; $L$ is the set of links; the indicator $\delta_l^p$ equals 1 if link $l$ is in path $p$ and equals 0 otherwise; $f_l$ is traffic flow on link $l$; $w_l$ is the toll on link $l$; function $C_l(\cdot)$ maps the link flows to link costs; $T_p(\cdot)$ is a non-linear function that maps the sum of link tolls to a cost value that is regarded as part of the path cost; and $\mu$ is a toll adjustment factor.

Using these settings, the performance of our models (IRL-F and CRL-F) and the ZM model are again compared under the nine trajectory sampling rates (Scenarios C1–C9) and the resulting WAPE values are shown in Figure 8. The ZM model produces much higher estimation errors compared to the results in scenario set B. Our models outperform the ZM model in all scenarios with different trajectory sampling rates and prior OD demands. The results from scenario set C are important because it is common that certain behavioural assumptions used in traffic assignment and simulation models may not fully reflect the real-world behaviours. This highlights the advantage of the proposed data-driven approach over traditional simulation-based approaches in inferring the population travel patterns by effectively leveraging the information embedded in the available data without relying on prior knowledge or behavioural assumptions.

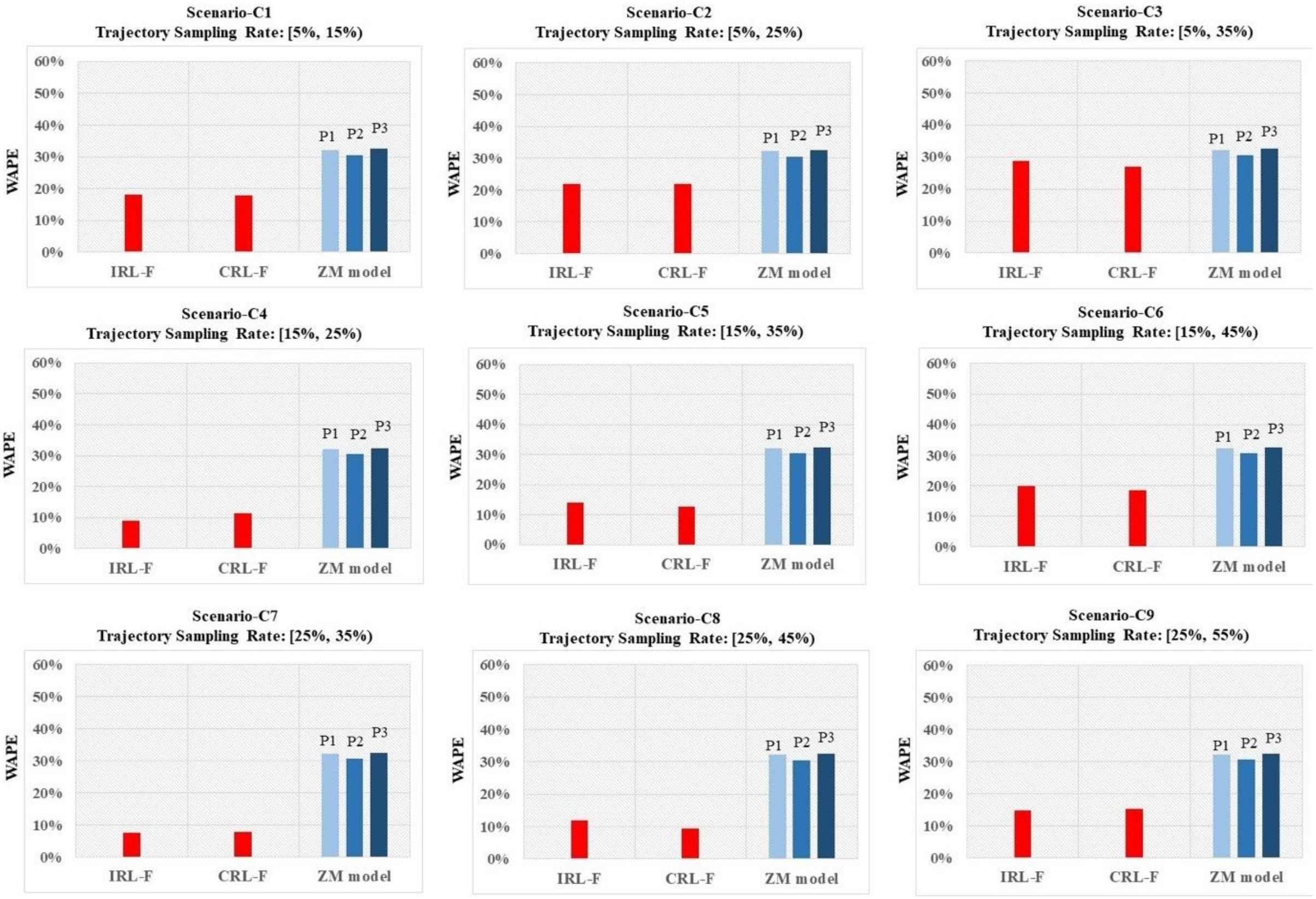

Figure 8. Performance comparison with ZM model for scenario set C

### 4.2.2 Comparison with the BHR (Brunauer-Henneberger-Rehrl) model

Two sets of test scenarios are designed to compare our proposed model to the BHR model. Scenario set D satisfies the trajectory data coverage assumed by the BHR model, while scenario set E relaxes this assumption.

**Scenario set D:** It is assumed that the observed vehicle trajectories cover most link-to-link transitions in the road network so that the propagation rules defined in the BHR model are valid. To create this scenario set, traffic assignment is first conducted on the Nguyen-Dupuis network under user equilibrium conditions to generate the ground-truth path set. Initially, the set of paths assigned non-zero flows does not reach a high enough coverage of the link-to-link transitions. Thus, we randomly select some unused paths, assign them random path flows, and add these paths to the ground-truth path set. This makes the adjusted path set cover about 80% of the link transitions on the network. The link flow estimation is performed under the nine different trajectory sampling rates (Scenarios D1–D9). The comparison of WAPE between our models (IRL-F and CRL-F) and the BHR model is shown in Figure 9. The performance of IRL-F and CRL-F varies with the trajectory sampling rate especially with the width of its range, while the BHR model shows less variations across different trajectory sampling rates. When the range of sampling rate is larger (e.g., Scenarios D3, D6, and D9), the BHR model shows comparable or slightly better performance than our models. When the range is smaller, however, our models show significantly better performance than the BHR model (e.g., Scenarios D1, D4, and D7).

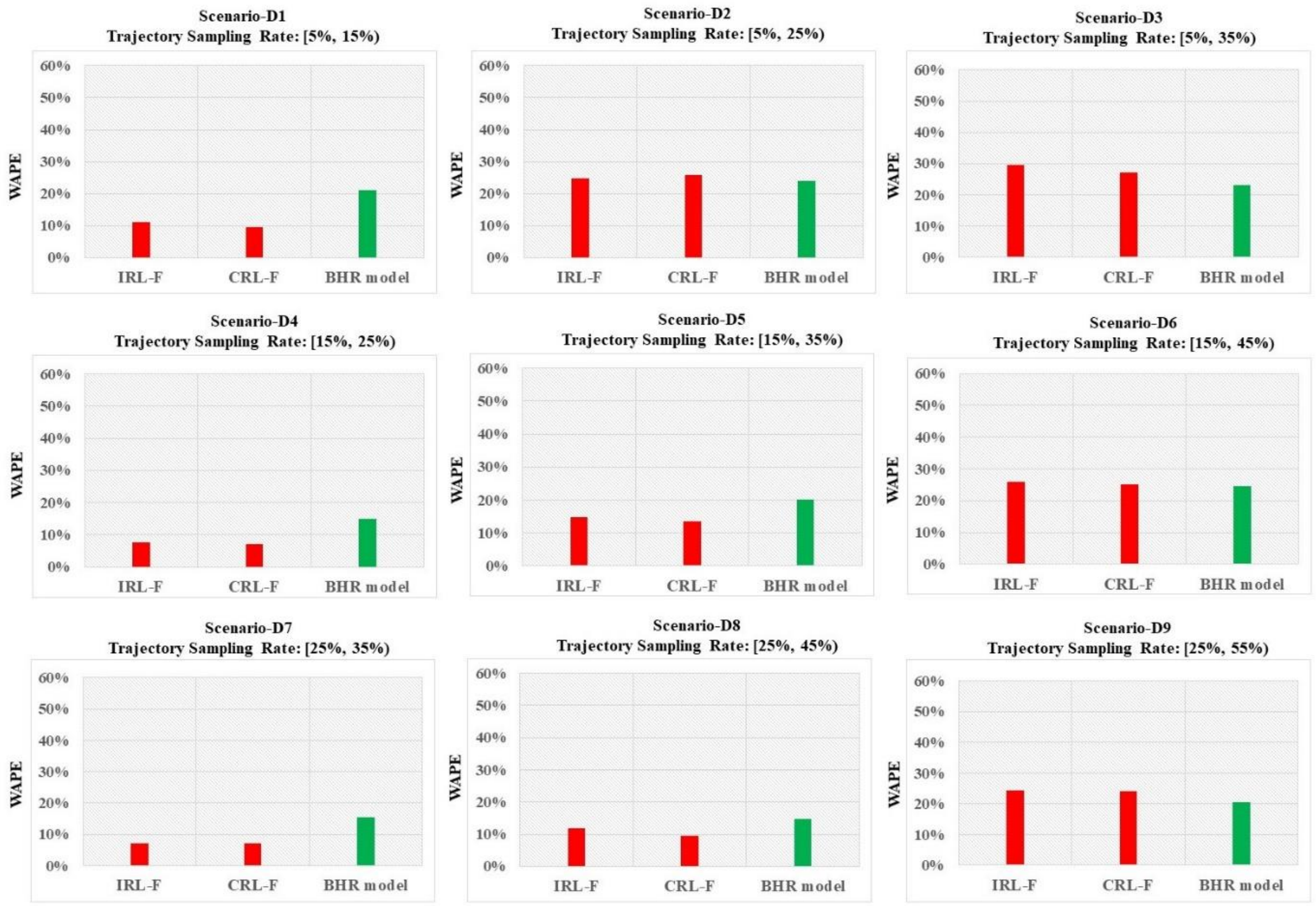

Figure 9. Performance comparison with BHR model for scenario set D

**Scenario set E:** We now relax the assumption on a high coverage of observed trajectory data to evaluate the models in a more realistic conditions as the penetration rates of available vehicle trajectory data are still quite low in many cities. To create such test environments, we only use the paths obtained by solving the traffic assignment problem on the Nguyen-Dupuis network as the ground-truth path set without further adjustment applied in scenario set D. This path set covers about 60% of the link-to-link transitions on the network, which is lower than the level of coverage used for scenario set D (i.e., 80%). The performance comparison under the nine trajectory sampling rates (Scenarios E1–E9) is shown in Figure 10. The estimation errors have significantly increased in the BHR model compared to scenario set D, suggesting that the propagation rules used in this method depend heavily on the coverage of observed trajectories and fail to estimate the true link flows when many of the link-to-link transitions have no observed trajectories. On the other hand, our models produce the similar performance to scenario set D, indicating a high level of robustness of the proposed generative approaches against the sparsity and low coverage of real-world trajectory data.

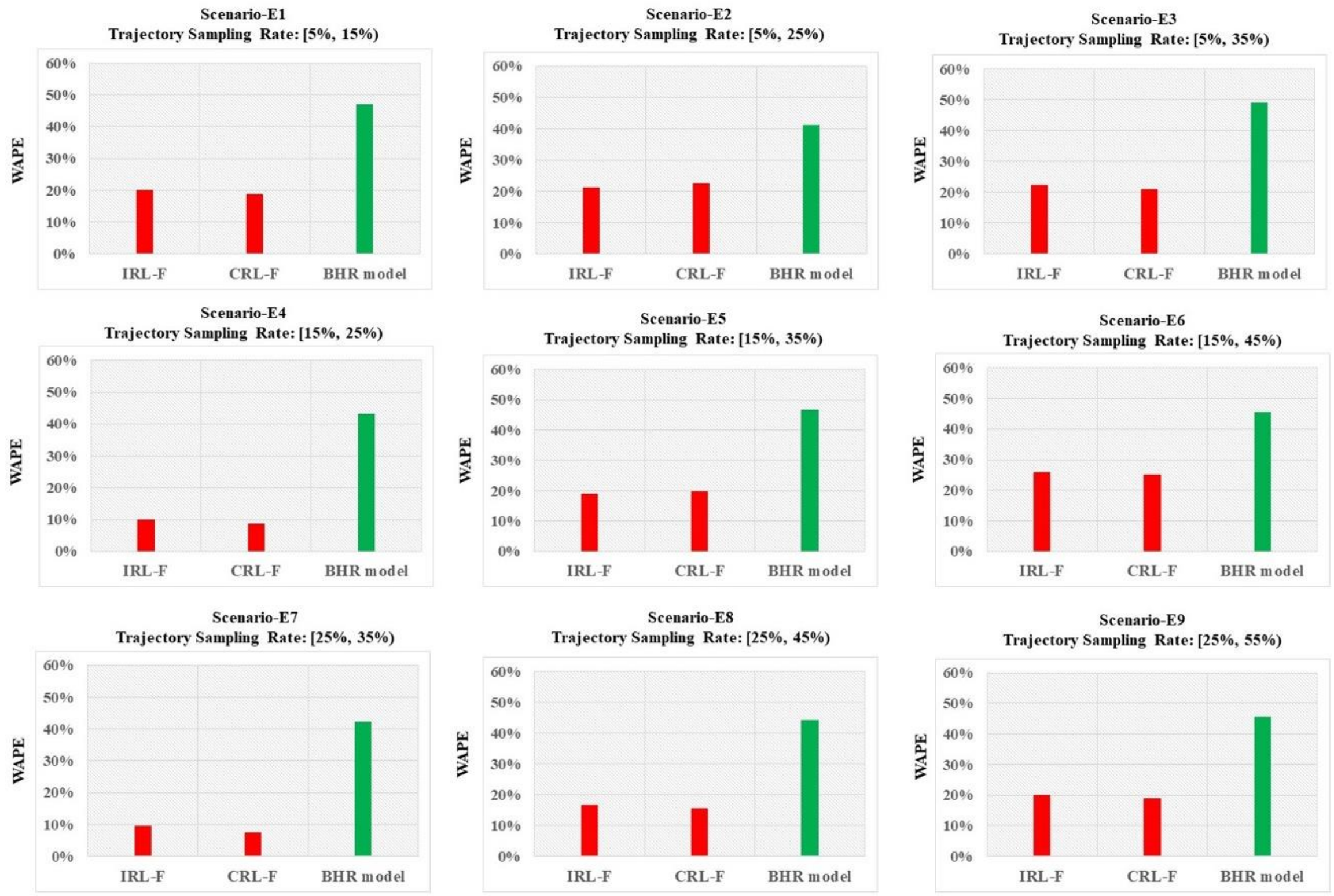

Figure 10. Performance comparison with BHR model for scenario set E

## 5. Conclusions

This paper proposes a novel data-driven approach to solving the link flow estimation problem when limited traffic volume data and sparse vehicle trajectory data are available. The main idea is to learn vehicle movement patterns from the available data and *generate synthetic population vehicle trajectories* to estimate link flows on unobserved links. This is different from the existing approaches that use traffic assignment models, which rely on strong behavioural assumptions, or that require stricter data availability such as a high spatial coverage of trajectory data. We develop a formal mathematical framework based on MDP that formulates the trajectory generation problem as a sequential decision problem in reinforcement learning (RL). In this RL framework, an agent (representing a vehicle) learns to make the link-to-link transition decisions across a road network based on a certain policy. We propose two RL-based solution approaches, namely IRL-F and CRL-F, to find the optimal policy that allows the synthetic trajectories generated by the agent to best match the available trajectory and traffic volume data. We extend the strategy of *matching feature expectations* between the real-world observed behaviour (from data or expert demonstrations) and the agent's behaviour (from model or policy) to solve our problem. More specifically, we propose new ways to define state feature vectors to allow both vehicle trajectory data and link traffic volume data to be used as expert demonstrations to generate synthetic population trajectories that conform to these observations.

Our generative modelling framework was validated using a real road network in Berlin, producing reasonable estimation results on test scenarios with different data availability settings. IRL-F and CRL-F performed similarly in all test scenarios, meaning that either of these methods can be applied to solve real-life link flow estimation problems. Using the Nguyen-Dupuis network, the proposed framework was compared to two methods from the literature under different driver behaviour and data availability settings. The proposed framework shows considerable advantage over the two benchmark methods especially under more realistic scenarios where certain behavioural assumptions about drivers are not met or the network coverage and penetration rate of trajectory data are low. Overall, our framework makes contributions to the link flow estimation applications by specifically dealing with challenging scenarios where modellers only have limited traffic volume data and sparse vehicle trajectory data but no prior knowledge about the travellers' demand or route choice behaviours. While this study focuses on the link flow estimation problem, the application of our framework is not limited to this problem. The synthetic population trajectories generated as the output of our models can be used to solve other traffic estimation problems such as estimating turning movements at intersections or travel demands for origin-destination pairs. Our framework can be viewed as a more general synthetic trajectory generator capable of generating realistic trajectories that would have caused the observed traffic counts and sample trajectory data. Such a trajectory generator has many applications in broader urban mobility studies including data augmentation (generating trajectories to complement existing sparse trajectory datasets), privacy protection (replacing privacy-sensitive real trajectories with synthetic trajectories), and mobility prediction (predicting individual or collective vehicle movement patterns). In the future work, our framework can be further improved in several directions. We plan to consider a more sophisticated method to determine the scaling factor to scale up generated trajectories to obtain the population trajectories. The computational efficiency is also an important issue, which we plan to improve by adopting more efficient RL algorithms or advanced deep RL/IRL architectures. For applications beyond the link flow estimation problem, the model evaluation should consider more diverse metrics to assess the quality of individual generated trajectories and their representativeness of the true population trajectories.